\pdfoutput=1

\documentclass[11pt]{article}

\usepackage[final]{acl}

\usepackage{times}
\usepackage{latexsym}
\usepackage{multirow}
\usepackage{algorithm}
\usepackage{algorithmic}

\usepackage[T1]{fontenc}

\usepackage[utf8]{inputenc}

\usepackage{microtype}

\usepackage{inconsolata}

\usepackage{graphicx}

\title{RAEmoLLM: Retrieval Augmented LLMs for Cross-Domain Misinformation Detection Using In-Context Learning \\ Based on Emotional Information}

\author{%
  Zhiwei Liu\textsuperscript{1} \quad
  Kailai Yang\textsuperscript{1} \quad 
  \textbf{Qianqian Xie}\textsuperscript{2} \quad 
  \textbf{Christine de Kock}\textsuperscript{\textbf{3}} \\ 
  \textbf{Sophia Ananiadou}\textsuperscript{\textbf{1}}\textsuperscript{\thanks{Corresponding Author}} \quad 
  \textbf{Eduard Hovy}\textsuperscript{\textbf{3},\textbf{4}}\\
    \textsuperscript{1} The University of Manchester \quad 
    \textsuperscript{2} School of Artificial Intelligence, Wuhan University \quad \\
    \textsuperscript{3} The University of Melbourne \quad  
    \textsuperscript{4} Carnegie Mellon University \\
\texttt{\{zhiwei.liu,kailai.yang,sophia.ananiadou\}@manchester.ac.uk}\\
\texttt{xqq.sincere@gmail.com,} \quad \texttt{\{christine.dekock,eduard.hovy\}@unimelb.edu.au}\\
}

\begin{document}
\maketitle
\begin{abstract}
Misinformation is prevalent in various fields such as education, politics, health, etc., causing significant harm to society. However, current methods for cross-domain misinformation detection rely on effort- and resource-intensive fine-tuning and complex model structures. With the outstanding performance of LLMs, many studies have employed them for misinformation detection. Unfortunately, they focus on in-domain tasks and do not incorporate significant sentiment and emotion features (which we jointly call {\em affect}). In this paper, we propose RAEmoLLM, the first retrieval augmented (RAG) LLMs framework to address cross-domain misinformation detection using in-context learning based on affective information. 
RAEmoLLM includes three modules. (1) In the index construction module, we apply an emotional LLM to obtain affective embeddings from all domains to construct a retrieval database. (2) The retrieval module uses the database to recommend top K examples (text-label pairs) from source domain data for target domain contents. (3) These examples are adopted as few-shot demonstrations for the inference module to process the target domain content. The RAEmoLLM can effectively enhance the general performance of LLMs in cross-domain misinformation detection tasks through affect-based retrieval, without fine-tuning. We evaluate our framework on three misinformation benchmarks. Results show that RAEmoLLM achieves significant improvements compared to the other few-shot methods on three datasets, with the highest increases of 15.64\%, 31.18\%, and 15.73\% respectively. This project is available at https://github.com/lzw108/RAEmoLLM.
\end{abstract}

\section{Introduction}

The internet is flooded with misinformation \cite{scheufele2019science2}, which has a significant impact on people's lives and societal stability \cite{della2023misinformation5}. Misinformation is pervasive across various domains such as education, health, technology, and especially on the internet, which requires people to invest significant time and effort in discerning the truth \cite{perez2018automatic1}. However, models trained in specific known domains are often fragile and prone to making incorrect predictions when presented with samples from new domains \cite{saikh2020deep}. As a result, detecting cross-domain misinformation has become an urgent global issue and poses greater challenges and difficulties.

Although some studies address cross-domain misinformation detection \cite{comito2023towards4,tang2023learning1,shi2023rough2}, they require effort-intensive fine-tuning, and apply only traditional machine learning methods or complex deep learning methods. Recently, LLMs have achieved impressive results in various tasks through zero-shot, few-shot \cite{li2023practical3}, or instruction tuning \cite{zhang2023instruction2}. Many researchers have applied LLMs to identify misinformation \cite{li2023self1,hu2024bad2,cheung2023factllama}. 
However, these methods perform only in-domain misinformation detection. Moreover, emotions and sentiments (which we jointly call {\em affect}) are important characteristics of human expression and communication \cite{hakak2017emotion}. When authors publish misinformation, they often consciously choose specific emotions to capture the attention and resonance of readers to encourage rapid spread \cite{keen2006theory,liu2024emotion2}. Unfortunately, there are few LLMs that utilize affective information to detect misinformation, and the only ConspEmoLLM \cite{liu2024conspemollm3} are developed based on an emotional LLM, which does not make full use of affective information, has no cross-domain ability, and also needs time-consuming fine-tuning.

In-context learning (ICL) needs only task instructions and few-shot examples (input-label pairs), eliminating fine-tuning on specific task labels \cite{dong2022survey3}. A few studies have used ICL to address cross-domain problems \cite{long2023adapt5,wu2024domain4}. To the best of our knowledge, there is currently no application of ICL for cross-domain misinformation detection based on affective information retrieval.

To address these issues, we propose the first retrieval augmented (RAG) LLMs framework based on emotional information (RAEmoLLM), to address cross-domain misinformation detection using in-context learning based on affective information.
RAEmoLLM contains three modules: (1) In the \textit{index construction} module, we apply EmoLLaMA-chat-7B \cite{liu2024emollms} to encode all domain corpora, obtaining implicit affective embeddings to construct the retrieval database as well as explicit affective labels. We also conduct a comprehensive affective analysis to demonstrate the effectiveness of affective information for discriminating between true and misinformation. (2) The \textit{retrieval} module recommends the top K affect-related examples (text-label pairs) from the source domain corpus according to the target domain content,  obtained from the retrieval database. (3) These examples are utilized as the few-shot demonstrations in the \textit{inference} module, which is driven by a prompt template to guide the LLM to verify the target content for misinformation. The template helps combine implicit and explicit affective information. This framework effectively enhances the capabilities of LLMs in multiple cross-domain misinformation detection tasks through leveraging affective information, without the need for fine-tuning.

\noindent 
In this work, we make three main contributions:
\begin{itemize} 
\item We conduct affective analysis on different kinds of misinformation datasets and construct the retrieval database according to the implicit affective information for misinformation datasets.
\item We propose RAEmoLLM, the first framework for cross-domain misinformation detection using ICL based on affective information, which does not require fine-tuning. Experimental results show that RAEmoLLM outperforms the zero-shot method and other few-shot methods. 
\item We evaluate RAEmoLLM on a variety of misinformation benchmarks, including fake news, rumours, and conspiracy theory datasets. Results show that RAEmoLLM achieves significant improvements compared to the other few-shot methods on three datasets, with the highest increases of 15.64\%, 31.18\%, and 15.73\% respectively, which illustrate the effectiveness of RAEmoLLM framework.
\end{itemize}

\section{Methodology}

\begin{figure*}[htb]
\centering
\includegraphics[width=1.8\columnwidth]{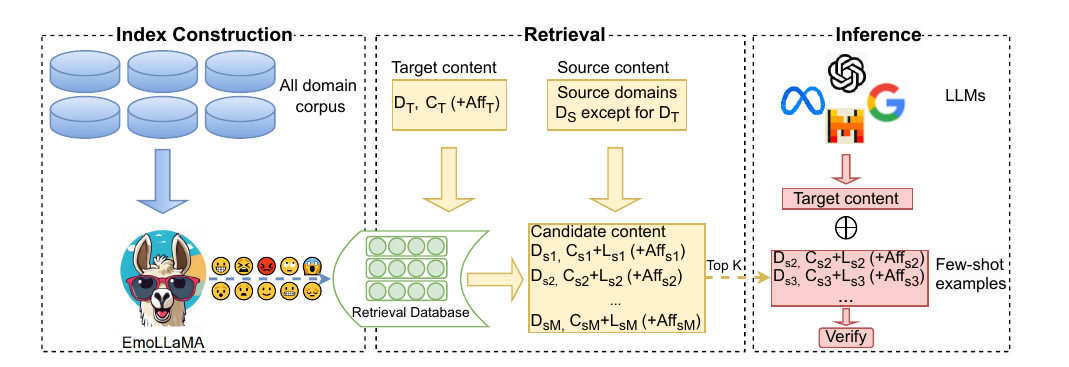}
\caption{The architecture of RAEmoLLM. D: Domain. T: Target domain. S: Source domain. C: Corpus. L: Label. Aff: Affective information. M: Number of source domain data. Index Construction Module: Apply an emotional LLM to obtain affective embeddings to construct a retrieval database. Retrieval Module: Recommend top K examples (text-label pairs) from source domain data. Inference Module: Adopt the recommended examples as demonstrations for inference.}
\label{fig:RAEmoLLM}
\end{figure*}

This section introduces our method of cross-domain misinformation detection, using the \textit{index construction} module, \textit{retrieval} module and \textit{inference} module. 
The overall architecture of RAEmoLLM is shown in Figure~\ref{fig:RAEmoLLM}. In the \textit{index construction} module (Sec. \ref{sec:indexconstructmodule}), we collect domain datasets, and employ an emotional LLM to obtain affective embeddings as well as affective labels to conduct a comprehensive affective analysis on them to detect the affective differences between real and false information. The implicit embeddings are adopted to construct the retrieval database, which will be used by the \textit{retrieval} module (Sec. \ref{sec:retrievalmodule}) to obtain source-domain examples.  These results are used as the few-shot examples for \textit{inference} module's (Sec. \ref{sec:inferencemodule}) in-context learning to detect target domain misinformation.

\subsection{Index Construction Module \label{sec:indexconstructmodule}}

In this section, we first introduce the original datasets and the processing procedure at Sec. \ref{subsec:originaldatasets}. We subsequently conduct affective analysis on these datasets and present how and why to obtain implicit and explicit affective information at Sec. \ref{subsec:aff5dimonsions}. Finally, we apply the implicit affective information to construct the retrieval database (Sec. \ref{subsec:retrievaldatabase}).

\subsubsection{Datasets \label{subsec:originaldatasets}}

\begin{table*}[htb]
\footnotesize
\resizebox{\textwidth}{!}{
\begin{tabular}{lcc|lcc|lcc}
\hline
\multicolumn{3}{c|}{AMTCele}                                 & \multicolumn{3}{c|}{PHEME}                & \multicolumn{3}{c}{COCO}                                        \\
Domain        & Legit                & Fake                  & Events            & Rumours & Non-rumours & Topics            & Related              & Conspiracy           \\ \hline
Technology    & 40                   & 40                    & Charlie Hebdo     & 458     & 1621        & Fake Virus        & \multirow{3}{*}{248} & \multirow{3}{*}{612} \\
Education     & 40                   & 40                    & Sydney siege      & 522     & 699         & Harmful Radiation &                      &                      \\
Business      & 40                   & 40                    & Ferguson          & 284     & 859         & Depopulation      &                      &                      \\
Sports        & 40                   & 40                    & Ottawa shooting   & 470     & 420         & Other 9 domains            & 540                  & 1181                 \\
Politics      & 40                   & 40                    & Germanwings-crash & 238     & 231         & Total             & 788                  & 1793                 \\
Entertainment & 40                   & 40                    & Putin missing     & 126     & 112         &                   & \multicolumn{1}{l}{} & \multicolumn{1}{l}{} \\
Celebrities   & 250                  & 250                   & Prince Toronto    & 229     & 4           &                   & \multicolumn{1}{l}{} & \multicolumn{1}{l}{} \\
Total         & 490                  & 490                   & Gurlitt           & 61      & 77          &                   & \multicolumn{1}{l}{} & \multicolumn{1}{l}{} \\
              & \multicolumn{1}{l}{} & \multicolumn{1}{l|}{} & Ebola Essien      & 14      & 0           &                   & \multicolumn{1}{l}{} & \multicolumn{1}{l}{} \\
              & \multicolumn{1}{l}{} & \multicolumn{1}{l|}{} & Total             & 2402    & 4023        &                   & \multicolumn{1}{l}{} & \multicolumn{1}{l}{} \\ \hline
\end{tabular}
}
\caption{\label{tab:datasetsthree}
Statistics of datasets. AMTCele includes 7 domains. PHEME contains 9 domains (events). COCO has 12 domains (topics). For AMTCele and PHEME, we apply leave-one-domain-out strategy for evaluation. For COCO, we select 3 domains as test set.}
\end{table*}

\begin{table*}[!t]
\footnotesize
\centering
\begin{tabular}{ccccccccc}
\hline
\multirow{2}{*}{Datasets} & \multirow{2}{*}{Affective} & \multirow{2}{*}{sub-emotion} & \multicolumn{2}{c}{legit/non-rumours/related} & \multicolumn{2}{c}{fake/rumours/conspiracy} & \multicolumn{2}{c}{t-test} \\
                          &                            &                              & mean                  & var                   & mean                 & var                  & t            & p           \\ \hline
\multirow{5}{*}{AMTCele}  & \multirow{4}{*}{EIreg}     & Anger                        & 0.3584                & 0.0064                & 0.4055               & 0.0060               & -9.3294      & 6.91E-20    \\
                          &                            & Fear                         & 0.3587                & 0.0137                & 0.4047               & 0.0124               & -6.2861      & 4.90E-10    \\
                          &                            & Joy                          & 0.3392                & 0.0180                & 0.2897               & 0.0142               & 6.1054       & 1.48E-09    \\
                          &                            & Sadness                      & 0.3341                & 0.0109                & 0.3697               & 0.0106               & -5.3726      & 9.70E-08    \\
                          & Vreg                       & -         & 0.5471                & 0.0204                & 0.4940               & 0.0170               & 6.0656       & 1.88E-09    \\ 
PHEME                     & EIreg                      & Sadness                      & 0.5215                & 0.0152                & 0.5177               & 0.0182               & 1.1442       & 0.2526      \\
COCO                      & Vreg                       & -                             & 0.3961                & 0.0095                & 0.3973               & 0.0066               & -0.3325      & 0.7395      \\ \hline
\end{tabular}
\caption{\label{tab:statisticsvaluesreg}
Statistics values of EIreg and Vreg on different datasets. The t-test is conducted between \textit{legit/non-rumours/related} and \textit{fake/rumours/conspiracy}. The complete statistics on PHEME and COCO can be found in Table \ref{tab:statisticsvaluesreg2} in the Appendix \ref{app:affanalysis}.}
\end{table*}

We collect FakeNewsAMT \cite{perez2018automatic1}, Celebrity \cite{perez2018automatic1}, PHEME \cite{kochkina2018all2}, and COCO \cite{langguth2023coco3} datasets. The statistics of these datasets are presented in Table \ref{tab:datasetsthree}. 
FakeNewsAMT is a cross-domain dataset, including six domains. The legitimate news in FakeNewsAMT was obtained from various mainstream news websites. The authors adopted crowdsourcing via Amazon Mechanical Turk (AMT) to generate fake versions of legitimate news items. The Celebrity dataset was derived from online magazines. We combine FakeNewsAMT and Celebrity as AMTCele. PHEME contains a collection of Twitter rumours and non-rumours posted during nine breaking news events. 
COCO dataset consists of 12 conspiracy theory categories\footnote{Suppressed Cures, Behavior Control, Anti Vaccination, Fake Virus, Intentional Pandemic, Harmful Radiation, Depopulation, New World Order, Esoteric Misinformation, Satanism, Other Conspiracy Theory, Other Misinformation.}. 
Each tweet in COCO is assigned an overall intention label, as follows: \textit{Conspiracy} is assigned to tweets for which the tweet is related to at least one of the 12 categories and is actively spreading conspiracy theories. Otherwise, if the tweet is related to the specific category, but it does not propagate misinformation or conspiracy theories, then the overall label of \textit{Related} is used. The overall label of \textit{Unrelated} is only used for tweets that are unrelated to all 12 conspiracy categories. We remove the \textit{Unrelated} text since the aim of the cross-domain test. 

For AMTCele and PHEME, we apply leave-one-domain-out strategy\footnote{By sequentially selecting a specific domain as the test set and the remaining domains as the training set, we can evaluate the model's performance on each individual domain and combine these results to obtain a comprehensive assessment of the overall dataset.} to evaluate the model. For COCO dataset, due to one text data may involve one or multiple topics, we select all data points involving the \textit{Fake Virus}, \textit{Harmful Radiation}, and \textit{Depopulation} topics as the test set, and the other topics as the retrieval dataset.

\subsubsection{Affective Analysis \label{subsec:aff5dimonsions}}

We firstly conduct a comprehensive affective analysis after collecting datasets. EmoLLaMA-chat-7B, which has the best overall performance among the EmoLLMs \cite{liu2024emollms}, is used for affective analysis. EmoLLaMA-chat-7B can be used to extract five kinds of affective dimensions (which we jointly call affect), including Emotion intensity (EIreg), Emotion intensity classification (EIoc), Sentiment (valence) strength (Vreg), Sentiment (valence) classification (Voc) and Emotion detection (Ec). The detailed introduction can be found in Appendix \ref{app:fiveaffanalysis}.


\textbf{Obtain implicit and explicit affective information:} Following the guidelines of EmoLLMs \cite{liu2024emollms}, we add prompts provided by EmoLLMs for each data point in order to obtain vectors from the last hidden layer (i.e., 4096d) for each affective dimension, as well as final labels using EmoLLaMA-chat-7B. We subsequently determine the distribution of affective information in different categories in each dataset. 

\textbf{Explicit affective analysis:} Table \ref{tab:statisticsvaluesreg} and Table \ref{tab:statisticsvaluesreg2} show regression information (i.e., EIreg and Vreg) of final labels. We use the t-test\footnote{t-test is a statistical method used to compare whether the difference between the means of two sets of data is significant. It generates a t-value, which is then compared to a t-distribution to determine if the observed difference is significant.} to measure the difference in emotional intensity between two sets of data. The t-value and p-value calculated between \textit{legit/non-rumours/related} and \textit{fake/rumours/conspiracy} demonstrate that there are statistically significant affective differences between the different categories. Figure \ref{fig:EIocAMT} to Figure \ref{fig:EcPHEME} and the chi-squared test in Appendix \ref{app:furtheraffectiveanalysis} confirm that other classifications using affective information are also related to misinformation. However, Table \ref{tab:statisticsvaluesreg} also presents some special cases that cannot effectively distinguish real and false information (e.g. EIreg-sadness in PHEME, Vreg in COCO). \citet{liu2024conspemollm3} also conducted some experiments that demonstrated that simply utilizing explicit affective information does not enhance the model's capability. Therefore, we introduce implicit affective information.

\textbf{Implicit affective analysis:} Table \ref{tab:statisticsvaluesEmbeddings2} shows statistics of different affective embeddings (i.e. last hidden layer of EmoLLaMA-chat-7B). We perform t-tests on the top-$K$ cosine similarity within categories and across categories. The results indicate that the similarity within categories is significantly higher than across categories, confirming that similar top-$K$ data points are likely to belong to the same category (further analysis can be found in Appendix \ref{app:furtheraffectiveanalysis}). We also visualize the data distribution reduced to 3 dimensions using PCA in Figures \ref{fig:3DvisualizationAMT1} and \ref{fig:3DvisualizationPHEME1} in Appendix. It can be observed that different categories are clearly separated in the latent space. All the above demonstrate the close relationship between affective information and misinformation.

\subsubsection{Retrieval Database Construction \label{subsec:retrievaldatabase}}



After obtaining the implicit affective embeddings in the previous step, we proceed to construct a comprehensive retrieval database. This database consists of vectors that encapsulate rich affective information, enabling efficient retrieval and analysis.

\subsection{Retrieval Module \label{sec:retrievalmodule}}

\begin{algorithm} 
\footnotesize
	\caption{Retrieval process} 
	\label{alg1} 
	\begin{algorithmic}[1]
		\REQUIRE Target domain corpus $D_T$, source domain corpus $D_S$, retrieval database $R$. 
		\ENSURE Target domain corpus with top K retrieval examples $D_{retri}$. 
		\STATE $E_T \gets R(D_T)$
            \STATE $E_S \gets R(D_S)$

            \FOR{$e_t$ in $E_T$}
            \FOR{$e_s$ in $E_S$}
            \STATE score = cosine($e_t$,$e_s$)
            \STATE $Sco \gets score$
            \ENDFOR
            \STATE $D_{retri} \gets$ select top k examples in $R(D_S)$ according to $Sco$
            \ENDFOR
	\end{algorithmic} 
\end{algorithm}

The retrieval database constructed in Sec \ref{sec:indexconstructmodule} is represented as R. Algorithm \ref{alg1} shows the retrieval process. In this module, we first process the multi-domain datasets into text-label pairs to obtain the target domain data $D_T=[\{c_{t1},l_{t1}\},\{c_{t2},l_{t2}\},...,\{c_{tN},l_{tN}\}]$ and source domain data $D_S=[\{c_{s1},l_{s1}\},\{c_{s2},l_{s2}\},...,\{c_{sM},l_{sM}\}]$ ($c$ denotes corpus text, and $l$ is the label. $N$ and $M$ are the numbers of target domain data and source domain data respectively). Following that, we obtain the target domain affective embedding $E_T=[e_{t1},e_{t2},...,e_{tN}]$ and source domain affective embedding $E_S=[e_{s1}, e_{s2},...,e_{sM}]$ through the embedding retrieval database R based on the corpus texts in $D_T$ and $D_S$.  Subsequently, we traverse the target domain embedding ($e_t$) in $E_T$ and calculate the similarity values with each source domain embedding $e_s$ from $E_S$ using the cosine method. Finally, we select the top k examples from source domain for each target domain data based on \textit{Sco} to $D_{retri}$, which will be the few-shot examples for LLM inference.

\subsection{Inference Module \label{sec:inferencemodule}}

We apply template 1 to construct the instruction datasets for inference once we get the top examples for each target domain data. \textit{[task prompt]} denotes the instruction for the task (The different \textit{[task prompts]} for each datasets can be found in Appendix \ref{app:taskpromptandexample}). \textit{[input text]} is a data item from the target domain data. \textit{[examples]} are the retrieval examples from source domain data (i.e. $D_{retri}$) and the \textit{[output]} is the output from LLM.

\begin{center}
\footnotesize
\fcolorbox{black}{gray!10}{
\begin{minipage}{0.45\textwidth}
\textbf{Template 1}  \\
\textbf{Task:} \textit{[task prompt]}  \\
\textbf{Target text:} \textit{[input text]}  \\
\textbf{Here are a few examples:} \textit{[examples]} \\
\textbf{According to the above information, the label of target text:} \textit{[output]}
\end{minipage}
}
\end{center}

We also apply template 2 to add explicit affective information. \textit{[affective information]} contains five dimensions described in Section \ref{subsec:aff5dimonsions}. The format of \textit{[examples]} is ``\textit{Text: [text]. [Affective info]: [value]. The label of text: [label]}''. Table \ref{tab:examplePHEME} shows one complete example.

\begin{center}
\footnotesize
\fcolorbox{black}{gray!10}{
\begin{minipage}{0.45\textwidth}
\textbf{Template 2}  \\
\textbf{Task:} \textit{[task prompt]}  \\
\textbf{Target text:} \textit{[input text]} + \textit{[affective info]} \\
\textbf{Here are a few examples retrieved by \textit{[affective info]}:} \textit{[examples]} \\
\textbf{According to the above information, the label of target text:} \textit{[output]}
\end{minipage}
}

\end{center}

\section{Experiments}




\begin{table*}[htb]
\resizebox{\textwidth}{!}{
\begin{tabular}{lcccccccccccc}
\hline
                         & \multicolumn{4}{c}{AMTCele}                                           & \multicolumn{4}{c}{PHEME}                                             & \multicolumn{4}{c}{COCO}                                              \\
Model                    & Acc             & Pre             & Rec             & F1              & Acc             & Pre             & Rec             & F1              & Acc             & Pre             & Rec             & F1              \\ \hline
BERT                     & 0.5414          & 0.5453          & 0.5414          & 0.5322          & 0.7214          & 0.7203          & 0.7214          & 0.7208          & 0.7288          & 0.7510          & 0.7288          & 0.6356          \\
RoBERTa                  & 0.5678          & 0.7228          & 0.5678          & 0.4730          & 0.7199          & 0.7213          & 0.7199          & 0.7204          & 0.7328          & 0.7851          & 0.7328          & 0.6388          \\
MDFEND                   & 0.5878          & 0.5934          & 0.5878          & 0.5815          & 0.5796          & 0.6425          & 0.5796          & 0.5829          & 0.7988          & 0.7939          & 0.7988          & 0.7793          \\
EDDFN                    & 0.7041          & 0.7313          & 0.7041          & 0.6951          & 0.7004          & 0.6925          & 0.7004          & 0.6816          & 0.7116          & 0.5064          & 0.7116          & 0.5917          \\
MOSE                     & 0.5031          & 0.5051          & 0.5031          & 0.4482          & 0.7135          & 0.7130          & 0.7135          & 0.6890          & 0.7198          & 0.7335          & 0.7198          & 0.6162          \\ 
CANMD           & 0.6296 & 0.6650 & 0.6296 & 0.6086 & 0.7382 & 0.7338 & 0.7382 & 0.7346 & 0.7291 & 0.7324 & 0.7291 & 0.6441 \\
MetaAdapt       & 0.6429 & 0.6564 & 0.6429 & 0.6350 & 0.6193 & 0.6804 & 0.6193 & 0.6230 & 0.5186 & 0.7267 & 0.5186 & 0.5222 \\ \hline
Mistral-7b-zs                 & 0.7020          & 0.7346          & 0.7020          & 0.6926          & 0.5897          & 0.6491          & 0.5897          & 0.5936          & 0.3686          & 0.7050          & 0.3686          & 0.4673          \\
Mistral-7b-random             & 0.7082          & 0.7768          & 0.7082          & 0.6889          & 0.6177          & 0.6334          & 0.6177          & 0.6227          & 0.7128          & 0.7455          & 0.7128          & 0.7287          \\
Mistral-7b-random-addexpl     & 0.6337          & 0.7050          & 0.6337          & 0.5988          & 0.5804          & 0.6177          & 0.5804          & 0.5870          & 0.6802          & 0.7245          & 0.6802          & 0.7010          \\
Mistral-7b-Vreg               & 0.7469          & 0.7748          & 0.7469          & 0.7404          & 0.6760          & 0.6837          & 0.6760          & 0.6788          & 0.7779          & 0.8031          & 0.7779          & 0.7898          \\
Mistral-7b-Vreg-addexpl       & \textbf{0.7735} & \textbf{0.7822} & \textbf{0.7735} & \textbf{0.7717} & \textbf{0.6921} & \textbf{0.6919} & \textbf{0.6921} & \textbf{0.6920} & \textbf{0.7814} & \textbf{0.8053} & \textbf{0.7814} & \textbf{0.7931} \\ \hline
Gemma-2b-zs                   & 0.4153          & 0.4568          & 0.4153          & 0.3815          & 0.3606          & 0.5113          & 0.3606          & 0.2303          & 0.3302          & 0.4572          & 0.3302          & 0.3835          \\
Gemma-2b-random               & 0.4980          & 0.4997          & 0.4980          & 0.4649          & 0.4269          & 0.5799          & 0.4269          & 0.3575          & 0.4477          & 0.6336          & 0.4477          & 0.4816          \\
Gemma-2b-random-addexpl       & 0.4929          & 0.4928          & 0.4929          & 0.4927          & \textbf{0.5914} & 0.5777          & \textbf{0.5914} & 0.5820          & 0.6221          & 0.6164          & 0.6221          & 0.5587          \\
Gemma-2b-Vreg                 & \textbf{0.6235} & \textbf{0.6298} & \textbf{0.6235} & \textbf{0.6213} & 0.4361          & \textbf{0.5953} & 0.4361          & 0.3708          & 0.5302          & \textbf{0.7326} & 0.5302          & 0.5814          \\
Gemma-2b-Vreg-addexpl         & 0.5847          & 0.6190          & 0.5847          & 0.5525          & 0.5875          & 0.5846          & 0.5875          & \textbf{0.5859} & \textbf{0.6767} & 0.6932          & \textbf{0.6767} & \textbf{0.5990} \\ \hline
Llama3.2-1b-zs             & 0.4796          & 0.4841          & 0.4796          & 0.4801          & 0.5549          & 0.4480          & 0.5549          & 0.4712          & 0.5826          & 0.5997          & 0.5826          & 0.5385          \\
Llama3.2-1b-random         & 0.5398          & 0.5483          & 0.5398          & 0.5222          & 0.3949          & 0.4831          & 0.3949          & 0.3417          & 0.7116          & 0.5064          & 0.7116          & 0.5917          \\
Llama3.2-1b-random-addexpl & 0.4867          & 0.4868          & 0.4867          & 0.4782          & 0.4118          & 0.4821          & 0.4118          & 0.3996          & 0.7116          & 0.5064          & 0.7116          & 0.5917          \\
Llama3.2-1b-Vreg           & 0.6173          & 0.6360          & 0.6173          & 0.6065          & 0.6254          & 0.6432          & 0.6254          & 0.6307          & 0.7233          & 0.7640          & 0.7233          & 0.6242          \\
Llama3.2-1b-Vreg-addexpl   & \textbf{0.6429} & \textbf{0.6460} & \textbf{0.6429} & \textbf{0.6438} & \textbf{0.6473} & \textbf{0.6831} & \textbf{0.6473} & \textbf{0.6535} & \textbf{0.7372} & \textbf{0.7718} & \textbf{0.7372} & \textbf{0.6545} \\ \hline
ChatGPT-zs                   & \textbf{0.7265} & 0.7420          & \textbf{0.7265} & \textbf{0.7221} & 0.5236          & 0.6551          & 0.5236          & 0.5032          & 0.7860          & 0.7920          & 0.7860          & 0.7551          \\
ChatGPT-random               & 0.6990          & 0.7475          & 0.6990          & 0.6835          & 0.6173          & 0.6539          & 0.6173          & 0.6234          & 0.7616          & 0.7782          & 0.7616          & 0.7079          \\
ChatGPT-random-addexpl       & 0.6959          & 0.7193          & 0.6959          & 0.6876          & 0.6092          & 0.6584          & 0.6092          & 0.6144          & 0.7651          & 0.7824          & 0.7651          & 0.7174          \\
ChatGPT-Vreg                 & 0.6745          & 0.7366          & 0.6745          & 0.6516          & \textbf{0.6370} & 0.6681          & \textbf{0.6370} & \textbf{0.6429} & \textbf{0.8151} & \textbf{0.8249} & \textbf{0.8151} & \textbf{0.7925} \\
ChatGPT-Vreg-addexpl         & 0.7163          & \textbf{0.7628} & 0.7163          & 0.7032          & 0.6318          & \textbf{0.6762} & 0.6318          & 0.6372          & 0.8012          & 0.8068          & 0.8012          & 0.7772          \\ \hline
GPT4o-zs                     & 0.8816          & 0.8856          & 0.8816          & 0.8813          & 0.6170          & 0.6398          & 0.6170          & 0.6228          & 0.7837          & 0.8150          & 0.7837          & 0.7396          \\
GPT4o-random                 & 0.8776          & 0.8850          & 0.8776          & 0.8770          & 0.6739          & 0.6830          & 0.6739          & 0.6771          & 0.8291          & 0.8526          & 0.8291          & 0.8090          \\
GPT4o-random-addexpl         & 0.8724          & 0.8824          & 0.8724          & 0.8716          & 0.6559          & 0.6693          & 0.6559          & 0.6601          & 0.8337          & 0.8527          & 0.8337          & 0.8158          \\
GPT4o-Vreg                   & \textbf{0.8888} & \textbf{0.8934} & \textbf{0.8888} & \textbf{0.8884} & 0.7004          & 0.6983          & 0.7004          & 0.6992          & \textbf{0.8477} & \textbf{0.8627} & \textbf{0.8477} & \textbf{0.8326} \\
GPT4o-Vreg-addexpl           & 0.8847          & 0.8912          & 0.8847          & 0.8842          & \textbf{0.7155} & \textbf{0.7170} & \textbf{0.7155} & \textbf{0.7162} & 0.8419          & 0.8605          & 0.8419          & 0.8242          \\ \hline
Vicuna-7b-zs                 & 0.5490          & 0.5545          & 0.5490          & 0.5384          & 0.4378          & 0.6502          & 0.4378          & 0.3542          & 0.2942          & 0.7054          & 0.2942          & 0.1592          \\
Vicuna-7b-random             & 0.5837          & 0.5872          & 0.5837          & 0.5806          & 0.4073          & 0.6116          & 0.4073          & 0.3017          & 0.7070          & 0.6037          & 0.7070          & 0.5928          \\
Vicuna-7b-random-addexpl     & 0.5622          & 0.6040          & 0.5622          & 0.5206          & 0.5334          & 0.5849          & 0.5334          & 0.5423          & 0.7023          & 0.5063          & 0.7023          & 0.5884          \\
Vicuna-7b-Vreg               & 0.6000          & 0.6069          & 0.6000          & 0.6023          & 0.4512          & \textbf{0.6549} & 0.4512          & 0.3821          & \textbf{0.7837} & \textbf{0.7999} & \textbf{0.7837} & 0.7471          \\
Vicuna-7b-Vreg-addexpl       & \textbf{0.6316} & \textbf{0.6680} & \textbf{0.6316} & \textbf{0.6248} & \textbf{0.6065} & 0.6145          & \textbf{0.6065} & \textbf{0.6105} & 0.7756          & 0.7956          & 0.7756          & \textbf{0.7501} \\ \hline
\end{tabular}}
\caption{\label{tab:resultsthree}
Overall results on three datasets. ``zs'' denotes the zero-shot method. ``random'' denotes randomly sample four examples without using affective information. ``random-addexpl'' denotes adding explicit Vreg information for the random sample examples. ``Vreg'' denotes retrieving four examples based on Vreg information using Template 1. ``Vreg-addexpl'' denotes adding explicit Vreg information using Template 2.}
\end{table*}

\subsection{Base Models}

\begin{itemize} 
\footnotesize

\item \textbf{LLMs:} We apply ChatGPT (gpt-3.5-turbo-0125), GPT-4o\footnote{https://openai.com/}, Llama3-8b-Instruct, Llama3.2-(1b,3b)-Instruct\footnote{https://llama.meta.com/llama3/}, Gemma-instruct-(2b, 7b) \cite{team2024gemma1}, Mistral-7b-Instruct \cite{jiang2023mistral4} and Vicuna-(7b, 13b, 33b) \cite{chiang2023vicuna2} as  base models to test our methods. 

\vspace{-1mm}

\item \textbf{PLMs:} We select BERT \cite{devlin2018bert6} and RoBERTa \cite{liu2019roberta7} as fine-tuning baselines. Specifically, one domain is selected as the target domain, other domains are used as the training dataset to fine-tune. 

\vspace{-1mm}

\item \textbf{Domain generalization methods (DGMs):} MOSE \cite{qin2020multitask} is a multi-domain mixture-of-experts (MoE) model, and each domain has its specific head. EDDFN \cite{silva2021embracing} preserves domain-specific and domain-shared knowledge. MDFEND \cite{nan2021mdfend} utilizes a Domain Gate to select useful experts of MoE. CANMD \cite{yue2022contrastive} performs label shift correction and contrastive learning. MetaAdapt \cite{yue2023metaadapt} adopts a meta-learning approach for domain-adaptive few-shot misinformation detection.

\vspace{-1mm}

\item \textbf{Retrieval method according to other types of embeddings:} We use the last\_hidden\_state of RoBERTa and another popular sentiment model (i.e. Sentibert \cite{yin2020sentibert}) as semantic and another kind of sentiment representation of each sentence respectively, then apply the same process of RAEmoLLM to deploy the ablation experiment.

\vspace{-1mm}

\item \textbf{Zero-shot and few-shot methods:} We also develop experiments of zero-shot method (LLMs-zs), randomly sample examples without using affective information (LLMs-random), and randomly sample examples with explicit Vreg information (LLMs-random-addexpl) for baselines.

\end{itemize} 


\subsection{Evaluation Metric}

Misinformation detection is typically regarded as a classification task, therefore we employ a variety of metrics—Accuracy, Precision, Recall, and F1 for evaluation \cite{su2020motivations3} (All metrics use the weighted variant).

\subsection{Results}

We evaluate RAEmoLLM framework on one Nvidia Tesla A100 GPU with 80GB of memory. The max length of new tokens is 256 and do\_sample is False. Others all use the default setting in the ``model.generate''\footnote{https://huggingface.co/docs/transformers/en/main\_classes/\\text\_generation} package. We firstly select the instruction data based on Vreg to test the effectiveness of the RAEmoLLM framework on different LLMs.  
The result is the overall performance, which means that in AMTCele and PHEME, every domain is considered as the target domain test set, and the overall result is the performance of the combination of each domain test set. For Gemma series, Llama series and Vicuna series, we only show the best overall performing one in the table. In this section, we will be discussing results exclusively based on the F1 score. We firstly compare the RAEmoLLM framework with various baseline methods (e.g. PLMs, domain generalization methods, zero-shot, and few-shot methods) at Sec. \ref{subsec:comparewithbaselines}. The ablation study of each module is conducted at Sec. \ref{subsec:ablation}. We subsequently compare the results on the data retrieved based on different affective information at Sec. \ref{subsec:comparedifaff}.

\subsubsection{Comparison with baselines \label{subsec:comparewithbaselines}}

\textbf{(1) Comparison with PLMs and other domain generalization methods:} We can observe that most LLMs with RAEmoLLM framework outperform fine-tuned RoBERTa, BERT, and DGMs on AMTCele and COCO datasets, but they slightly underperform fine-tuned models and some DGMs in the PHEME dataset. One possible reason is that in cross-domain misinformation detection tasks, the fine-tuning method may perform better for simple short-text discrimination problems in the large-scale dataset (e.g. PHEME). However, they may not be suitable for long texts (e.g. AMTCele) or complex tasks (e.g. intent recognition in COCO), especially in small datasets. We can see that the current DGMs do not have stable performance on different datasets, although they have complex structures. And their results are lower than the best performance of LLMs with the RAEmoLLM framework in most cases.

\textbf{(2) Comparison with zero-shot method (LLMs-zs), random few-shot methods (LLMs-random, LLMs-random-addexpl):} From Table \ref{tab:resultsthree}, we can observe that the RAEmoLLM framework largely increases the LLMs with zero-shot method in most cases and performs better than the random few-shot methods (For random few-shot, the largest increase in AMTCele is Gemma-2b (+15.64\%), in PHEME is Llama3.2-1b (+31.18\%), and in COCO is Vicuna-7b (+15.73\%)). The results of LLMs-random-addexpl show that simply applying explicit information has little effect in most cases\footnote{For Llama3.2-1b in COCO, both the random and random-addexpl variants predict all items as conspiracy category, resulting in the same results.}. 
A special case is that in the AMTCele dataset, GPT-4o and ChatGPT perform well in zero-shot settings, with ChatGPT even surpassing other few-shot methods. One possible reason is that the AMTCele dataset is collected from fact-checking websites, and ChatGPT's and GPT-4o's training set includes the data and can effectively utilize this information. One example is shown in Table \ref{tab:dataleakageexample}.

Table \ref{tab:differentdomainsAMTCele} and Table \ref{tab:differentdomainsPHEME} in Appendix \ref{appendix:resultsondifferentdomains} present the performance of Mistral-7b on each domain on AMTCele and PHEME separately. It can be observed that Mistral-7b with RAEmoLLM framework overtakes Mistral with zero-shot and few-shot methods in most domains except for the prince domain in PHEME, which has significantly imbalanced data. Additionally, we also conduct some special cases analysis in Appendix \ref{sec:special}.


\subsubsection{Ablation analysis of each module \label{subsec:ablation}}

\textbf{(1) Index Construction Module (retrieval based on different information):} From Table \ref{tab:resultsthree}, we can observe retrieval based on affective information (LLMs-Vreg, LLMs-Vreg-addexpl) overtake non-retrieval methods (i.e. random few-shot methods (LLMs-random, LLMs-random-addexpl)). From Table \ref{tab:differentemb}, we can observe that the RAEmoLLM framework achieves the best results compared to other types of embeddings, which indicates the effectiveness of Vreg embedding. 

\begin{table}[htb]
\footnotesize
\begin{tabular}{lccc}
\hline
{\color[HTML]{212121} }   & AMT    & PHEME  & COCO   \\ \hline
Mistral-7b-Vreg          & 0.7404 & 0.6788 & 0.7898 \\
Mistral-7b-Vreg-addexpl  & 0.7717 & 0.6920 & 0.7931 \\
Mistral-7b-semantic      & 0.6904 & 0.6718 & 0.7771 \\
Mistral-7b-sentibert     & 0.6984 & 0.6663 & 0.7687 \\ 
\hline
\end{tabular}
\caption{\label{tab:differentemb}
F1 score of retrieval using different kinds of embeddings. ``semantic'' denotes retrieval based on RoBERTa.}
\end{table}


\textbf{(2) Retrieval Module (different numbers of retrieval examples):}
Table \ref{tab:difnum} presents the F1 score of retrieval of different numbers of examples based on Vreg (we only tested 16 examples in the AMTCele dataset due to its long text). From the table, it can be observed that increasing the retrieval examples does not consistently improve the model's performance, and it may even lead to a decline in its performance (e.g. Vreg-addexpl in COCO). One possible reason is that when the model has multiple examples as references, it needs to consider a large amount of information comprehensively, which depends on the model's capability. Another reason we can infer from Table \ref{tab:statisticsvaluesEmbeddings2}. For the three datasets, the p-values in retrieval top 4 examples are all zero. However, as the number of retrieval examples increases, the second p-values in AMTCele and the first p-value in COCO dataset also gradually increase. This indicates that the retrieved content may come from another category or unrelated examples, thereby affecting the model's judgment ability. Therefore, when employing retrieval augmentation techniques, it is not just about blindly increasing the number of examples, but rather selectively choosing the most useful examples.

\begin{table}[!t]
\footnotesize
\resizebox{0.47\textwidth}{!}{
\begin{tabular}{llccccc}
\hline
Datasets                 & methods      & 4      & 8      & 16     & 32     & 64     \\ \hline
                                & Random       & 0.6889 & 0.7006 & 0.6287 & -      & -      \\
                                & Vreg         & 0.7404 & 0.7395 & 0.7271 & -      & -      \\
\multirow{-3}{*}{AMTCele}       & Vreg-addexpl & 0.7717 & 0.7611 & 0.7710 & -      & -      \\ \hline
                                & Random       & 0.6227 & 0.6253 & 0.6268 & 0.6400 & 0.6353 \\
                                & Vreg         & 0.6788 & 0.6856 & 0.6830 & 0.6910 & 0.7031 \\
\multirow{-3}{*}{PHEME}         & Vreg-addexpl & 0.6920 & 0.6949 & 0.6979 & 0.6979 & 0.6990 \\ \hline
                                & Random       & 0.7287 & 0.7534 & 0.7442 & 0.7628 & 0.7541 \\
                                & Vreg         & 0.7898 & 0.7842 & 0.7854 & 0.8172 & 0.7993 \\
\multirow{-3}{*}{COCO}          & Vreg-addexpl & 0.7931 & 0.7208 & 0.7499 & 0.7600 & 0.7475 \\ \hline
\end{tabular}
}
\caption{\label{tab:difnum}
F1 score of Mistral-7b with retrieval of different numbers of examples based on Vreg.}
\end{table}


\begin{figure*}[htb]
\centering
\includegraphics[width=2\columnwidth]{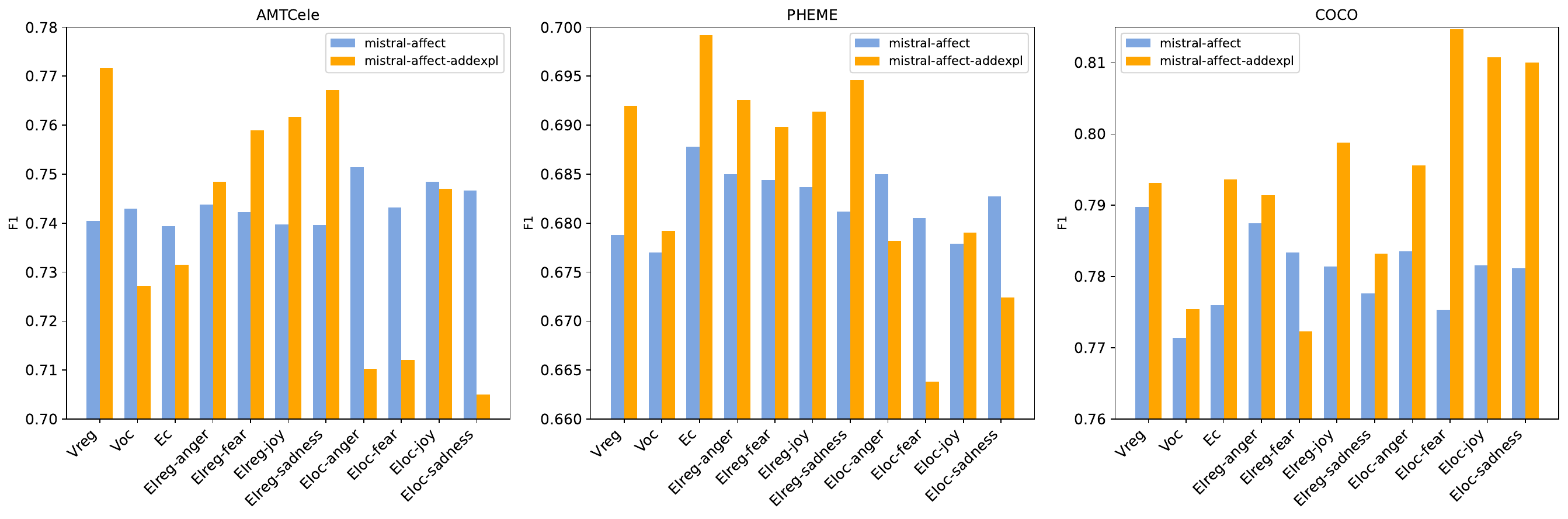}
\caption{Results of Mistral-7b based on different affective information on three datasets.  ``affect'' denotes retrieving four examples based on one affective information using Template 1. ``affect-addexpl'' denotes adding explicit affective information using Template 2.}
\label{fig:differentaffretri4_barfig}
\end{figure*}

\textbf{(3) Inference Module (different templates and different base LLMs):}
We can see LLMs with explicit affective information based on Template 2 (i.e. LLMs-Vreg-expl) exceed LLM without explicit affective information based on Template 1 (i.e. LLMs-Vreg) in most cases. For LLMs-zs and LLMs-random, different base models show significant performance differences. GPT-4o performs the best, followed by ChatGPT and Mistral-7b, while the Gemma-2b model has the lowest score. After using RAEmoLLM framework, the difference between different modules becomes narrowing (e.g. Mistral-7b has achieved or even surpassed the performance of ChatGPT.)

Based on the analysis above, we can conclude that retrieval based on implicit affective information and adding explicit affective information through Template 2 is the most effective way to enhance the LLMs' performance in using Vreg affective cases. The number of retrieval examples seems to have little impact. The LLMs focus on the most relevant examples.

Table \ref{tab:resultsthree} shows that Mistral-7b has the best performance among open-sourced LLMs. We choose Mistral-7b to conduct the following experiments.

\subsubsection{Results on the data retrieved based on different affective information \label{subsec:comparedifaff}}

Figure \ref{fig:differentaffretri4_barfig} presents the results of retrieval with different affective embeddings. For retrieval using affective regression information (i.e. Vreg, EIreg), it is evident that adding explicit affective information (\textit{affect-addexpl}) method can improve the performance compared to solely relying on implicit affective information retrieval (\textit{affect}). However, when using affective classification information (e.g. EIoc in AMTCele and PHEME), adding explicit affective information may confuse the model. In the COCO dataset,  all the \textit{affect-addexpl} method outperforms \textit{affect} except for EIreg-fear. Regarding the \textit{affect-addexpl} method, in AMTcele, we can see the results retrieval based on Vreg are best, followed by EIreg-sadness and EIreg-joy. And the final three rankings are retrieved based on EIoc-anger, fear, and sadness. It seems that affective intensity and strength are more suitable for cross-domain fake news detection tasks. In PHEME, retrieval based on Ec exhibits the highest performance, with the Vreg and EIreg series closely trailing behind. While the last few are the EIoc series, which may suggest that a coarse-grained emotional intensity classification is not suitable for rumour detection. However, it is the opposite in the conspiracy theory dataset. In COCO, the performance of retrieval based on the EIoc series is better than that based on the EIreg series. This could be attributed to the dataset's focus on COVID-19 topic, which may elicit more consistent emotional expressions among individuals.

\section{Conclusion and Future Work}

In this paper, we propose RAEmoLLM, the first RAG framework to address cross-domain misinformation detection using in-context learning based on affective information.  We introduce the three modules of RAEmoLLM. We also conduct a comprehensive affective analysis for three public misinformation datasets. We evaluate the performance of RAEmoLLM on the three misinformation benchmarks based on various LLMs. The results show that RAEmoLLM can significantly improve LLMs compared to the zero-shot method and other few-shot methods, which illustrates the effectiveness of RAEmoLLM. We also conduct an ablation analysis of each module and analyze the performance of retrieval based on different affective information, which provide a foundation for further improvements in the future.

In the future, we will explore the application of multimodal affective information in the task of detecting misinformation. We will also evaluate the application of the RAEmoLLM framework in other fields (e.g. mental health and finance). In addition to affective information, there are many other influencing factors in misinformation, such as stance and topic. We will combine sentiments and emotions with other features to construct a more robust retrieval database. Furthermore, the retrieval process can be slowed down by a large amount of data. In the future, we will also explore more efficient retrieval methods.

\section{Limitations}

Due to restricted computational resources, we only carried out inference of 1B, 2B, 7B, 8B, 13B, and 33B open-sourced LLMs. As such, we have not considered how the use of larger or different model architectures may potentially impact upon performance in cross-domain misinformation detection tasks.

Though achieving outstanding performance, RAEmoLLM still bears limitations. Firstly, for domain data with imbalanced distribution, RAEmoLLM performs worse compared to zero-shot methods (e.g. prince domain in PHEME). The special cases analysis in Appendix \ref{sec:special} also illustrates that in the imbalanced datasets, the retrieval in RAEmoLLM will be influenced for some special cases. Therefore, further exploration is needed to address such issues. Secondly, in the PHEME dataset, RAEmoLLM performs worse than fine-tuning methods without emotional information. This indicates that for simple tasks with shorter texts, the model still struggles to effectively balance textual features and emotional information.

\section*{Acknowledgments}

This work is supported by the Turing Scheme Fund, the Manchester-Melbourne-Toronto Research Fund, the Centre for Digital Trust and Society at the University of Manchester, and the New Energy and Industrial Technology Development Organization. This work is also supported by the scholar award from the Department of Computer Science at the University of Manchester.

\bibliography{acl_latex}


\appendix

\section{Related Work}

\subsection{Misinformation detection}

\textbf{Cross-domain misinformation detection:} Cross-domain misinformation detection refers to identifying and detecting misleading or false information across different domains or sources. \citet{comito2023towards4} propose a deep learning-based architecture to mitigate this problem by yielding high-level cross-domain features. \citet{tang2023learning1} design one framework to learn transferable features across domains by aligning the source and target news using Optimal Transport techniques. \citet{shi2023rough2} develop a rough-fuzzy graph learning framework that uses representations of cross-domain sample uncertainty structural information, and captures shared general features across domains. \citet{tong2024mmdfnd} integrate domain embeddings and attention mechanisms for domain-specific knowledge extraction and combine techniques to obtain multi-domain and multi-modal information.  \citet{nan2021mdfend} adopt domain gates to aggregate multiple representations extracted by a mixture of experts (MoE) for fake news detection. \citet{silva2021embracing} jointly leverage domain-specific and cross-domain knowledge and introduces an unsupervised technique to train a multi-domain fake news detection model. \citet{yue2022contrastive} propose a contrastive adaptation network, which leverages pseudo-labeling to generate target examples and design a label correction component to solve label shift problems. \citet{yue2023metaadapt} develop a domain-adaptive few-shot method based on meta-learning, which adopts limited target examples to provide feedback and guide knowledge transfer from the source domain to the target domain. However, these methods require complex structures and fine-tuning strategies.

\textbf{Retrieval augmented misinformation detection:} Retrieval augmented generation (RAG) combines LLMs with retrieval systems to utilize external knowledge, enabling models to generate more accurate content. \citet{xuan2024lemma} leverage LVLM intuition and reasoning capabilities to enhance the accuracy of multimodal misinformation by retrieving external knowledge. \citet{yue2024evidence} collect supporting evidence from scientific sources and generate responses for combating misinformation online based on this evidence. \citet{cheung2023factllama} adopt external, most up-to-date information available on the Internet to bridge the knowledge gap in an LLM to enhance fake news detection performance. \citet{li2024re} employ a multi-round retrieval strategy, which can extract key evidence from web sources for claim verification. These findings demonstrate the effectiveness of RAG technology in detecting misinformation.

\textbf{Affect-based misinformation detection:} Emotion and sentiment are important features for misinformation detection \cite{liu2024emotion2}. \citet{zhang2023sentiment1} combine the use of semantic and sentiment information, along with propagation information for rumour detection. \citet{dong2022sentiment2} design a sentiment-aware hyper-graph attention network for fake news detection. \citet{liu2024conspemollm3} develop a conspiracy theory detection LLM by fine-tuning EmoLLaMA \cite{liu2024emollms}. \citet{choudhry2022emotion5} utilize emotional information for fake news detection based on an adversarial learning structure. Unfortunately, these works either have complex structural designs or fine-tuned models, which require significant time and computational resources. RAEmoLLM in this paper applies the ICL method based on retrieving demonstration examples through affective information, which has a simple structure and does not involve fine-tuning.

\subsection{In-context learning}

In-context learning (ICL) is a specific prompting engineering method, in which the task demonstrations are included in prompts for LLMs learning \cite{xu2024context}. \citet{wang2023learning} develop a framework to provide high-quality context examples for LLMs, which firstly evaluate the quality of candidate examples through a reward model, and then conduct knowledge distillation to train a dense retriever. \citet{wang2024large} introduce an algorithm that utilizes a small LM to select the best demonstrations from a set of annotated data, and subsequently expand these demonstrations to larger LMs.  \citet{liu2024let1} develop in-context curriculum learning, a simple but helpful demonstration ordering method for ICL that gradually increases the complexity of prompt demonstrations. \citet{xu2024misconfidence2} propose in-context reflection to strategically select demonstrations that reduce the discrepancy between the LLM's outputs and the actual input-output mappings. \citet{long2023adapt5} propose a retrieval-enhanced language model to address cross-domain problems, in which they train language models by learning both target domain distribution and the discriminative task signal simultaneously with the augmented cross-domain in-context examples. Inspired by these works, we propose the RAEmoLLM.

\section{Task Prompt and Instruction Example \label{app:taskpromptandexample}}

For AMTCele, we utilize \textit{``Determine whether the target text is 0.~Fake or 1.~Legit.''} For PHEME, we employ \textit{``Determine if the target text is 0.~non-rumours or 1.~rumours.''} For COCO, we apply \textit{``Classify the text regarding COVID-19 conspiracy theories or misinformation into one of the following three classes: 0.~Unrelated. 1.~Related (but not supporting). 2.~Conspiracy (related and supporting).''} Here we keep the \textit{0.~Unrelated} category to test the robustness of the LLM by increasing the complexity of the task.

Table \ref{tab:examplePHEME} presents a specific instruction example.

\begin{table}[htb]
\footnotesize
\begin{tabular}{p{7.6cm}}
\hline
\textbf{Task}: Determine if the target text is 0. non-rumours or 1. rumours.                                                                                                                                     \\
\textbf{Target text}: UPDATE: Reports of 1 more shooter being SHOT. This is in addition to one shot and killed earlier in Parliament Hill \#OttawaShooting. Sentiment intensity: 0.234.                  \\
\textbf{Here are a few examples retrieved through sentiment intensity:}                                                                                                                                          \\
\textbf{Text}: UPDATE: Reports gunman says four devices are located around Sydney. Security response underway. Police calling for calm. \#9News. Sentiment intensity: 0.429. The label of this text: 1. rumours.  \\
\textbf{Text}: JUST IN: Police confirm to \@ABC there is a second hostage situation underway in eastern Paris. Sentiment intensity: 0.328. The label of this text: 1. rumours.                                    \\
\textbf{Text}: UPDATE: There are reports police have discovered the identity of the lone gunman, with the \#SydneySiege in its sixth hour. \#9News Sentiment intensity: 0.435. The label of this text: 1. rumours. \\
\textbf{Text}: JUST IN: A separate shooting and hostage situation at a supermarket in eastern Paris has been reported ... developing. Sentiment intensity: 0.236. The label of this text: 1. rumours.            \\
According to the above information, the label of target text:                                                                                                                                           \\ \hline
\end{tabular}
\caption{\label{tab:examplePHEME}
An example in the PHEME instruction dataset.}
\end{table}

\section{The results from different domains in the AMTCele and PHEME datasets. (Table \ref{tab:differentdomainsAMTCele} and \ref{tab:differentdomainsPHEME}) \label{appendix:resultsondifferentdomains}}

\begin{table*}[!t]
\footnotesize
\centering
\resizebox{\textwidth}{!}{
\begin{tabular}{lcccccccccccccc}
\hline
                      & \multicolumn{2}{c}{biz} & \multicolumn{2}{c}{edu} & \multicolumn{2}{c}{entmt} & \multicolumn{2}{c}{polit} & \multicolumn{2}{c}{sports} & \multicolumn{2}{c}{tech} & \multicolumn{2}{c}{celebrity} \\
Model                 & Acc        & F1         & Acc        & F1         & Acc         & F1          & Acc         & F1          & Acc          & F1          & Acc         & F1         & Acc           & F1            \\ \hline
BERT & 0.5975 & 0.5930 & 0.5725 & 0.5436 & 0.5800 & 0.5610 & 0.5450 & 0.5180 & 0.5525 & 0.5293 & 0.5650 & 0.5409 & 0.5152 & 0.5039        \\
Mistral-7b-zs                 & 0.7250 & 0.7135 & 0.8000 & 0.7954 & 0.7625 & 0.7595 & 0.5750 & 0.5157 & 0.7750 & 0.7714 & 0.6000 & 0.5442 & 0.6980 & 0.6925 \\
Mistral-7b-random             & 0.7375 & 0.7218 & 0.6625 & 0.6191 & 0.7375 & 0.7251 & 0.5500 & 0.4357 & 0.6875 & 0.6761 & 0.5625 & 0.4589 & 0.7580 & 0.7489 \\
Mistral-7b-Vreg               & 0.7750 & 0.7656 & 0.8250 & 0.8222 & 0.8250 & 0.8222 & 0.6125 & 0.5706 & 0.8125 & 0.8089 & 0.7250 & 0.7067 & 0.7320 & 0.7275 \\
Mistral-7b-Vreg-addexpl       & 0.8000 & 0.7968 & 0.8625 & 0.8620 & 0.8500 & 0.8496 & 0.6625 & 0.6423 & 0.8375 & 0.8373 & 0.8625 & 0.8607 & 0.7360 & 0.7346 \\ \hline
\end{tabular}
}
\caption{\label{tab:differentdomainsAMTCele}
The results from different domains in the AMTCele dataset}
\end{table*}

\begin{table*}[!t]
\resizebox{\textwidth}{!}{
\begin{tabular}{lcccccccccccccccccc}
\hline
                          & \multicolumn{2}{c}{sydneysiege} & \multicolumn{2}{c}{ottawashooting} & \multicolumn{2}{c}{charliehebdo} & \multicolumn{2}{c}{ferguson} & \multicolumn{2}{c}{germanwings} & \multicolumn{2}{c}{prince} & \multicolumn{2}{c}{putinmissing} & \multicolumn{2}{c}{gurlitt} & \multicolumn{2}{c}{ebola} \\
\multicolumn{1}{c}{Model} & Acc            & F1             & Acc              & F1              & Acc             & F1             & Acc           & F1           & Acc            & F1             & Acc          & F1          & Acc             & F1             & Acc          & F1           & Acc         & F1          \\ \hline
BERT & 0.7463 & 0.7418 & 0.7497 & 0.7490 & 0.7971 & 0.8113 & 0.7053 & 0.7147 & 0.7275 & 0.7260 & 0.1296 & 0.1985 & 0.5866 & 0.5297 & 0.5391 & 0.4949 & 0.5714 & 0.7220      \\
Mistral-7b-zs                 & 0.6536 & 0.6552 & 0.6506 & 0.6504 & 0.6075 & 0.6407 & 0.4051 & 0.4146 & 0.6716 & 0.6638 & 0.7382 & 0.8344 & 0.5546 & 0.4807 & 0.4420 & 0.4389 & 0.4286 & 0.6000 \\
Mistral-7b-random             & 0.6822 & 0.6838 & 0.5719 & 0.5232 & 0.6946 & 0.7153 & 0.4506 & 0.4653 & 0.6652 & 0.6646 & 0.6395 & 0.7636 & 0.5378 & 0.4569 & 0.5362 & 0.4225 & 0.3571 & 0.5263 \\
Mistral-7b-Vreg               & 0.7215 & 0.7195 & 0.6652 & 0.6596 & 0.7335 & 0.7521 & 0.5818 & 0.6102 & 0.7143 & 0.7139 & 0.5451 & 0.6881 & 0.6008 & 0.5716 & 0.4928 & 0.4514 & 0.5000 & 0.6667 \\
Mistral-7b-Vreg-addexpl       & 0.7437 & 0.7403 & 0.6753 & 0.6683 & 0.7431 & 0.7613 & 0.6527 & 0.6655 & 0.7036 & 0.7033 & 0.4592 & 0.6128 & 0.6050 & 0.6023 & 0.4348 & 0.4308 & 0.4286 & 0.6000 \\ \hline
\end{tabular}
}
\caption{\label{tab:differentdomainsPHEME}
The results from different domains in the PHEME dataset}
\end{table*}

\section{Special cases analysis\label{sec:special}}


\begin{table}[]
\centering
\resizebox{0.48\textwidth}{!}{
\begin{tabular}{llcccc}
\hline
\multirow{2}{*}{Datasets} & \multirow{2}{*}{EIoc} & \multirow{2}{*}{num} & \multirow{2}{*}{F1} & \multicolumn{2}{c}{mean num of retrieval} \\ \cline{5-6} 
                          &                       &                      &                     & legit/non-rum/related & fake/rumour/consp \\ \hline
                                                  & fake anger=0           & 218                   & 0.8152               & 1.0780                & 2.9220            \\
                                                  & legit anger=2/3        & 29                    & 0.9643               & 2.2414                & 1.7586            \\
                                                  & fake joy=2/3           & 14                    & 0.6667               & 1.5000                & 2.5000            \\
\multirow{-4}{*}{AMT}                             & legit joy=0            & 304                   & 0.8571               & 2.1217                & 1.8783            \\ \hline
                                                  & non-rum fear=2/3       & 446                   & 0.6978               & 2.4776                & 1.5224            \\
                                                  & rumour fear=0          & 1039                  & 0.3804               & 2.4658                & 1.5342            \\
                                                  & non-rum joy=0          & 3795                  & 0.8949               & 2.9057                & 1.0943            \\
\multirow{-4}{*}{PHEME}                           & rumour joy=2/3         & 25                    & 0.2759               & 3.7600                & 0.2400            \\ \hline
                                                  & related fear=2/3       & 47                    & 0.5538               & 2.0426                & 1.9574            \\
                                                  & consp fear=0           & 171                   & 0.9073               & 0.9708                & 3.0292            \\
\multirow{-3}{*}{COCO}                            & realted joy ==0        & 246                   & 0.7607               & 2.2927                & 1.7073            \\ \hline
\end{tabular}
}
\caption{\label{tab:specialcases} Special cases retrieval based on EIoc. ``num'' denotes number. ``non-rum'' denotes non-rumours. ``consp'' denotes conspiracy. The ``0'', ``2'', and ``3'' in the EIoc column represent ``no'', ``moderate'', and ``high'' emotional intensity.}
\end{table}

Misinformation and true information often convey different affective information (as shown in Table \ref{tab:statisticsvaluesreg} and Table \ref{tab:statisticsvaluesreg2}). For example, fake news and conspiracy theories tend to evoke more negative sentiments and emotions (e.g. anger or fear) and less joy. However, these results are based on statistics derived from the entire dataset. The special cases need to be analyzed. We investigate some special cases retrieved based on EIoc. The results are listed in Table \ref{tab:specialcases}.

For AMTCele, we investigate cases where fake news lacks anger or exhibits higher levels of joy, as well as cases where legit news displays higher levels of anger or lacks joy. We can see that the examples retrieved are mostly of the same category as the target, and their results have not been greatly influenced. For PHEME and COCO, we calculate statistics on cases of rumour and conspiracy without fear or exhibiting higher levels of joy (we do not report conspiracy with higher joy due to its low occurrence), as well as cases where non-rumour and related display higher levels of fear or without joy. We can see that the results for rumours in PHEME and related in COCO are poor. The most likely reason is due to the imbalance of categories in the original data, and these special cases are in the minority. This has resulted in the retrieval of more data from the larger category in original datasets, causing the model to learn less useful information and ultimately affecting the final results.

\section{Data leakage example in AMTCele \label{app:dataleakage} (Table \ref{tab:dataleakageexample})}

\begin{table}[]
\footnotesize
\begin{tabular}{p{1.2cm}p{5.8cm}}
\hline
Prompt & News content: Alex Jones Apologizes for Promoting 'Pizzagate' Hoax Alex Jones a prominent conspiracy theorist and the host of a popular right-wing radio show … \textbf{Where does this news come from?}                                                                                                                                                                                                                                                                                                         \\
GPT-3.5-trubo                 & This news comes from a New York Times article                                                                                                                                                                                                                                                                                                                                                                                                                                                                                                       \\
GPT-4o                        & The news about Alex Jones   apologizing for promoting the Pizzagate conspiracy theory likely comes from a   reputable news source such as The New York Times, The Washington Post, or   another major media outlet that covers significant events and developments involving   public figures and misinformation. The details provided in your question,   such as the mention of "The Alex Jones Show" and the Pizzagate   conspiracy, align with coverage typically found in mainstream news articles   addressing misinformation and its impact. \\ \hline
\end{tabular}
\caption{\label{tab:dataleakageexample}
Data leakage example in AMTCele}
\end{table}

\section{Comparison of time consumption between RAEmoLLM and fine-tuning methods \label{app:timeconsum} (Table \ref{tab:timeconsum})}

We take the PHEME dataset (6425 items) as an example to compare the time consumption between RAEmoLLM (applying ChatGPT as the base model) and the fine-tuning method (BERT). From Table \ref{tab:timeconsum}, it can be observed that RAEmoLLM will consume about 122s to construct the retrieval database (Obtain embeddings: 72s, Retrieval examples: 50s) and 208s to obtain the affective labels. For fine-tuning methods, we take BERT as an example. The current time consumed (Training each epoch: 3906s) by BERT was measured based on a single set of hyperparameters (e.g., batch size and learning rate). In practice, fine-tuning methods may require more time and effort to optimize hyperparameters. Overall, the RAEmoLLM process is simpler and more efficient. 

\begin{table*}[]
\centering
\footnotesize
\begin{tabular}{lllll}
\hline
                        & Obtain embeddings  & Obtain labels         & Retrieval Examples & Inference (time/item) \\
RAEmoLLM                & 71.68s             & 208s                  & 50s                & 0.48s                 \\ \hline
                        & Train (time/epoch) & Inference (time/item) &                    &                       \\
Bert                    & 3906.31s           & 0.093s                &                    &                       \\ \hline
\end{tabular}
\caption{\label{tab:timeconsum}
Time consumption of RAEmoLLM (take ChatGPT as base model) and fine-tuning methods (task BERT as the example) based on the PHEME dataset.}
\end{table*}

\section{Affective analysis \label{app:affanalysis}}

\subsection{Five types of affective information \label{app:fiveaffanalysis}}

(1) \textit{Emotion intensity (EIreg):} For each of four different emotions
(anger, fear, joy and sadness), assign a score between 0 and 1 to represent the intensity of emotion of the text; 

(2) \textit{Emotion intensity classification (EIoc):} The text can be classified into one of four classes of the intensity of emotion (anger, fear, joy, sadness), i.e. {no/low/moderate/high} emotional intensity; 

(3) \textit{Sentiment (valence) strength (Vreg):} Assign a real-valued score between 0 (most negative) and 1 (most positive) to represent the sentiment intensity of the text. 

(4) \textit{Sentiment (valence) classification (Voc):} The text can be categorized into one of seven ordinal classes (i.e. \textit{\{very, moderately, slightly\} negative, neutral, \{slightly, moderately, very\} positive}); 

(5) \textit{Emotion detection (Ec):} The text can be classified as ‘neutral or no emotion’ or as one, or more, of eleven given emotions (anger, anticipation, disgust, fear, joy, love, optimism, pessimism, sadness, surprise, trust).

\subsection{Further Affective Analysis \label{app:furtheraffectiveanalysis}}

\begin{figure}[htb]
\centering
\includegraphics[width=\columnwidth]{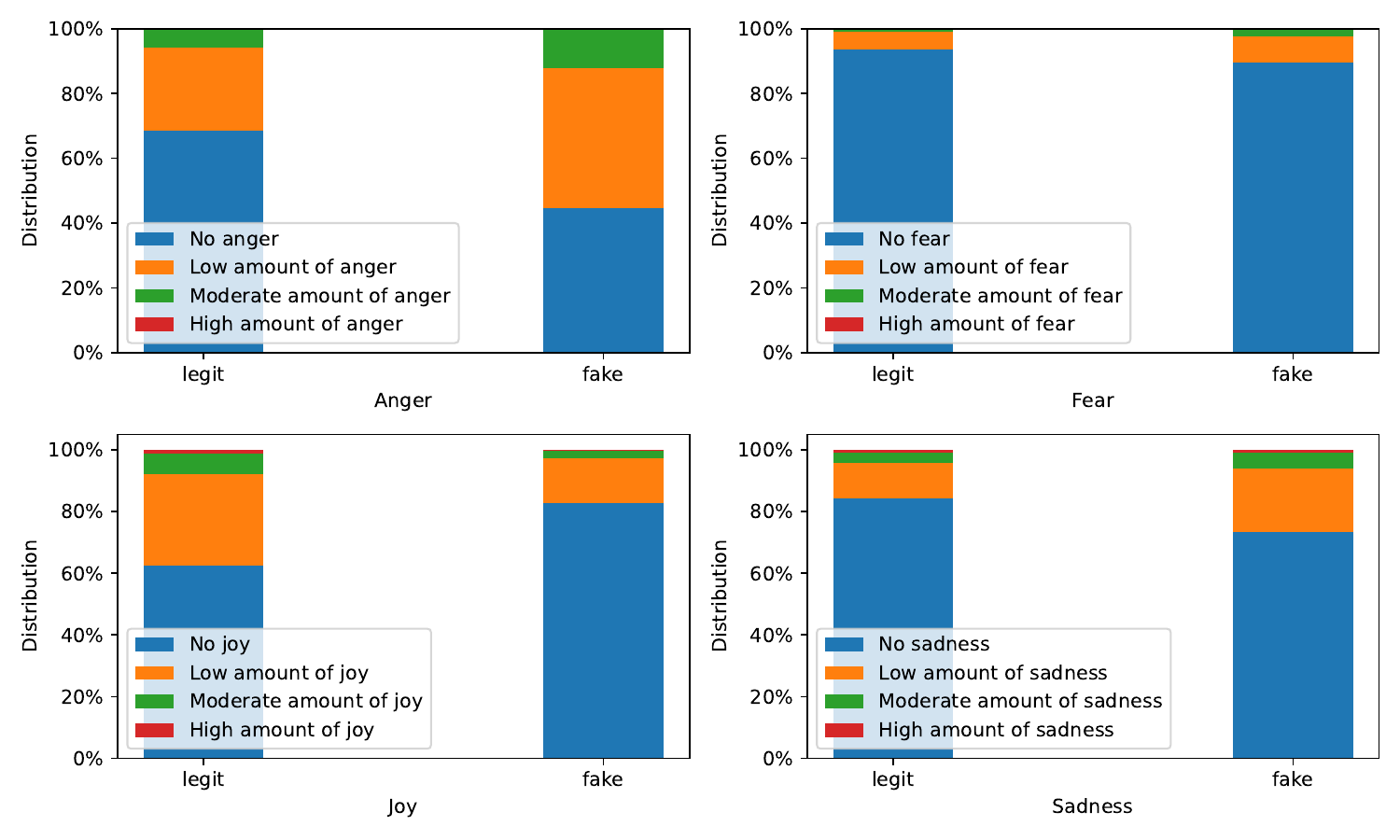}
\caption{Emotion intensity classification on AMTCele}
\label{fig:EIocAMT}
\end{figure}

\begin{figure}[htb]
\centering
\includegraphics[width=\columnwidth]{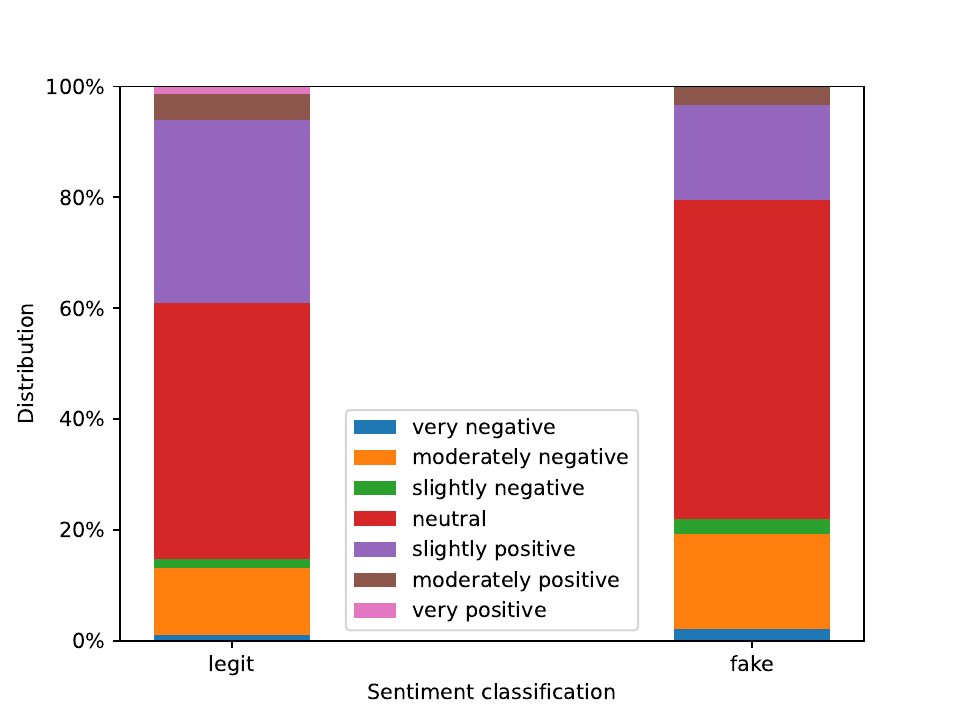}
\caption{Sentiment classification on AMTCele}
\label{fig:VocAMT}
\end{figure}

\begin{figure}[htb]
\centering
\includegraphics[width=\columnwidth]{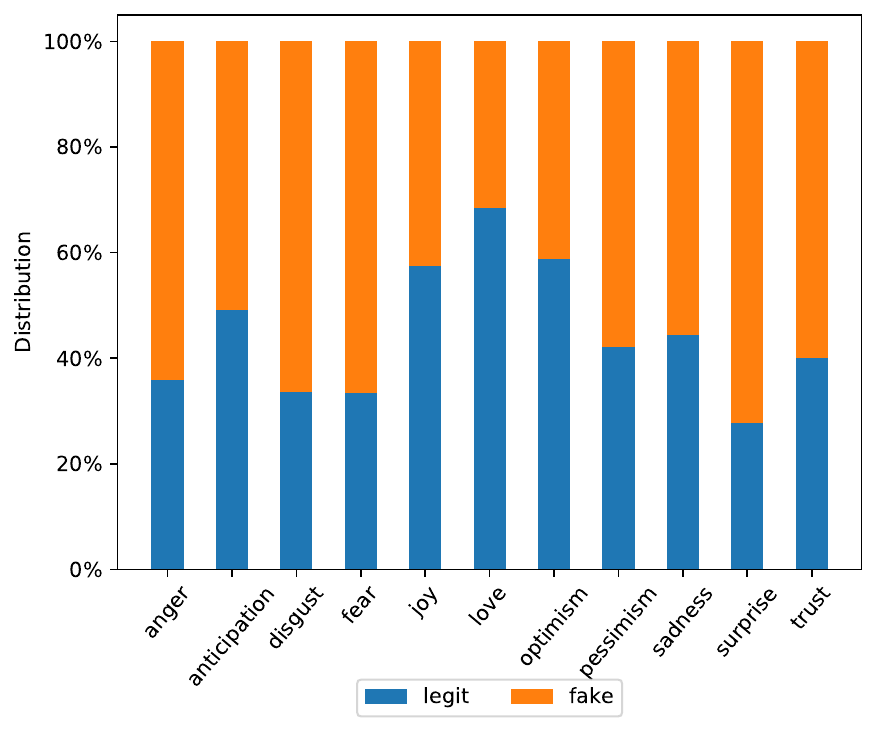}
\caption{Emotion classification on AMTCele}
\label{fig:EcAMT}
\end{figure}

\begin{figure}[htb]
\centering
\includegraphics[width=\columnwidth]{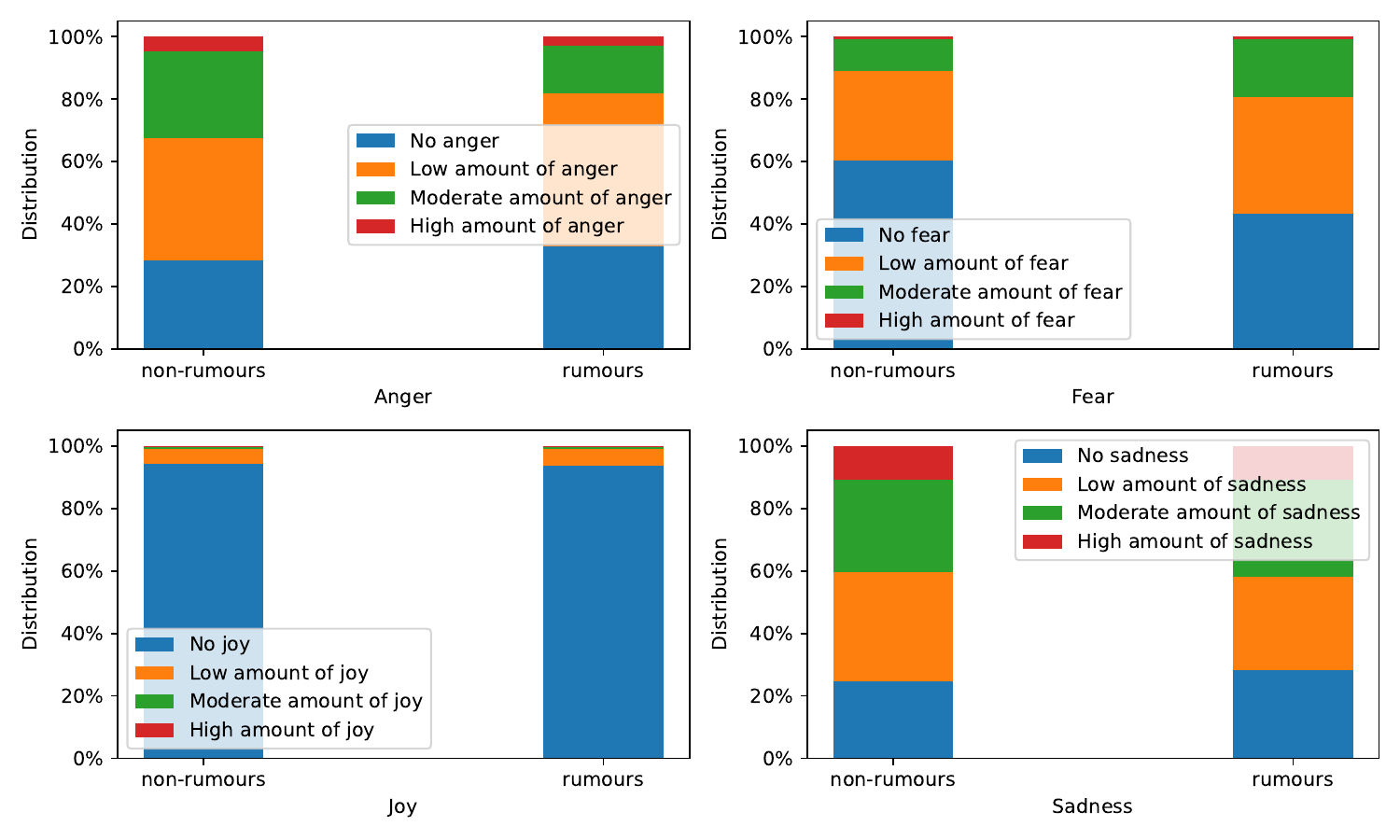}
\caption{Emotion intensity classification on PHEME}
\label{fig:EIocPHEME}
\end{figure}

\begin{figure}[htb]
\centering
\includegraphics[width=\columnwidth]{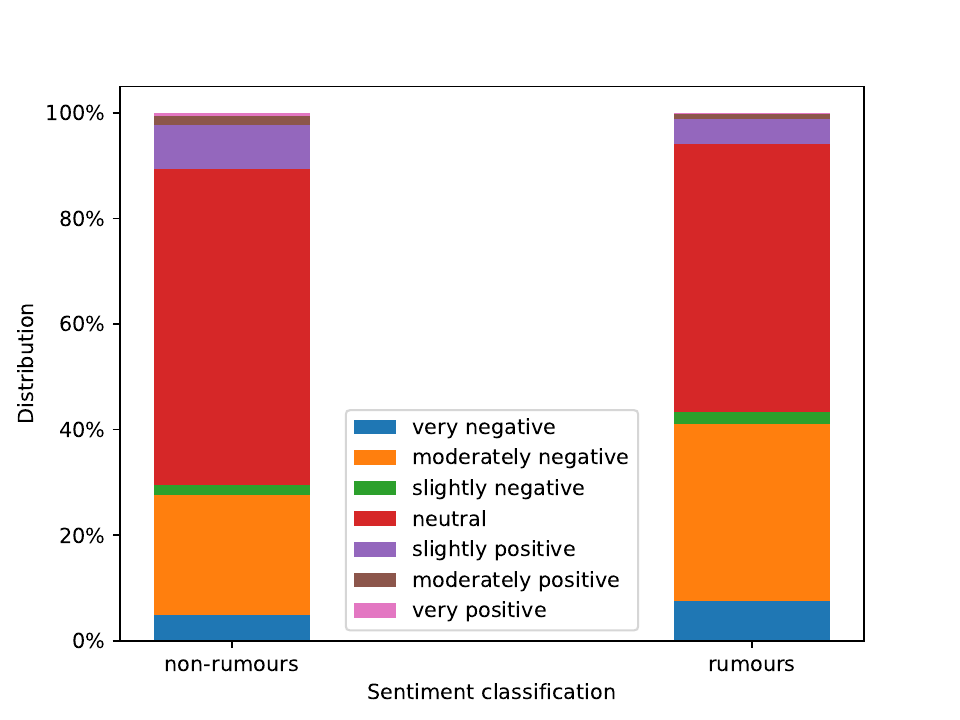}
\caption{Sentiment classification on PHEME}
\label{fig:VocPHEME}
\end{figure}

\begin{figure}[htb]
\centering
\includegraphics[width=\columnwidth]{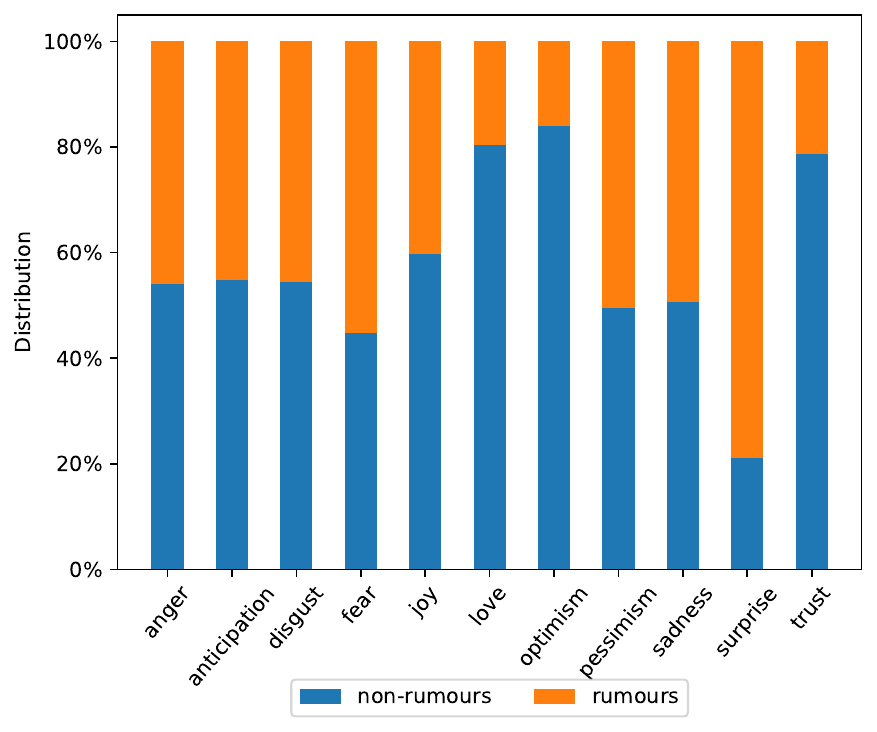}
\caption{Emotion classification on PHEME}
\label{fig:EcPHEME}
\end{figure}

\begin{figure*}[!t]
\centering
\includegraphics[width=1.7\columnwidth]{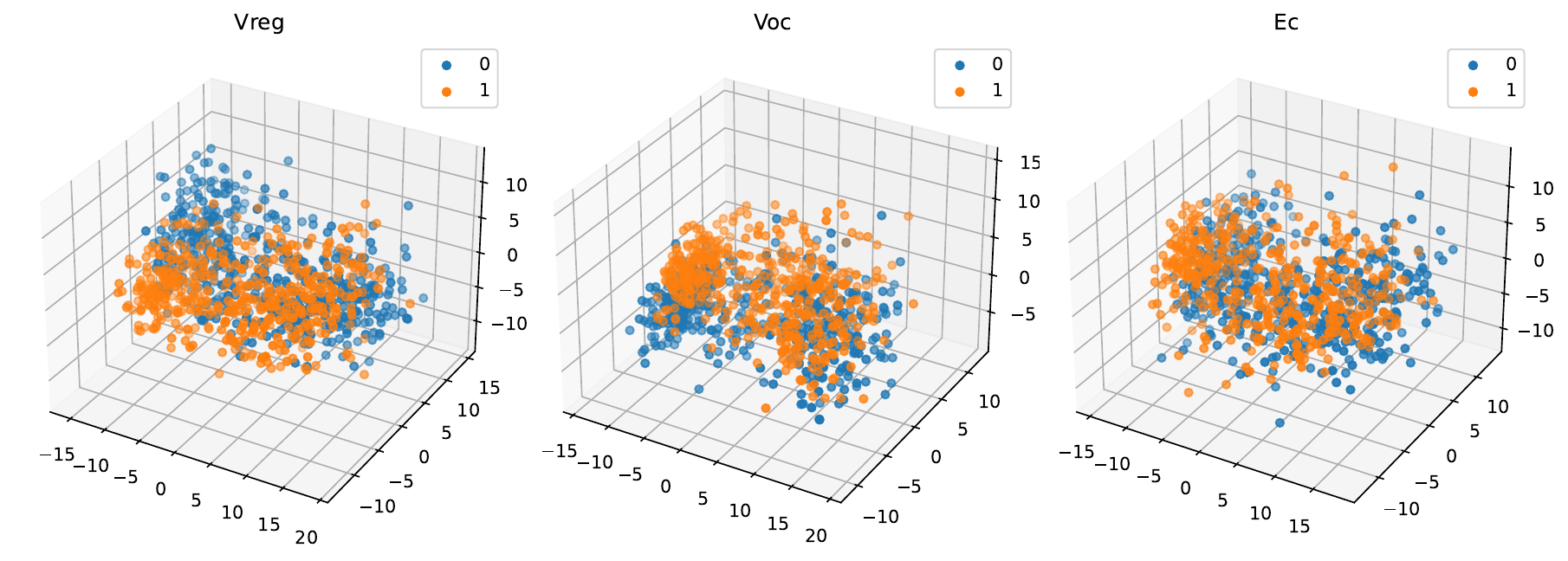}
\caption{3D visualization of affective embeddings on AMTCele. 0: Fake. 1: Legit}
\label{fig:3DvisualizationAMT1}
\end{figure*}

\begin{figure*}[!t]
\centering
\includegraphics[width=1.7\columnwidth]{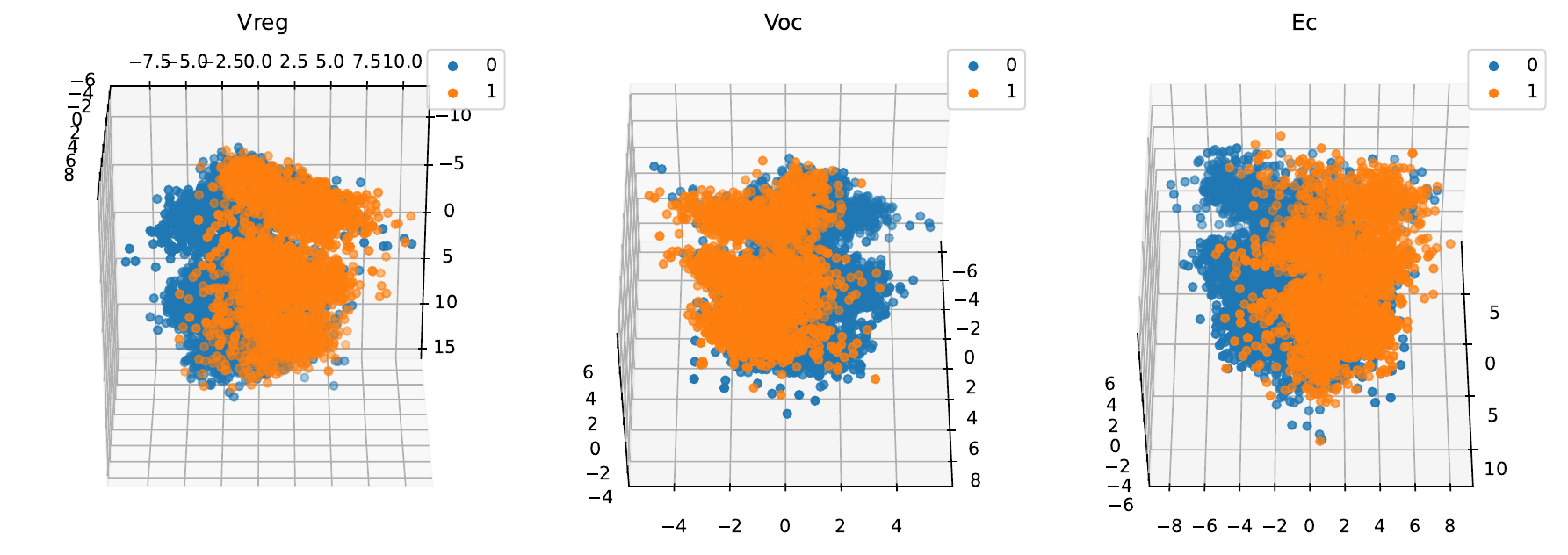}
\caption{3D visualization of affective embeddings on PHEME. 0: Non-rumours. 1: Rumours}
\label{fig:3DvisualizationPHEME1}
\end{figure*}

We show the statistics values and distribution of labels and embeddings in this Section. In Figures \ref{fig:EIocAMT} to Figure \ref{fig:EcPHEME}, the y-axis represents the distribution of labels within the intention class indicated on the x-axis. The affective analysis on COCO has been done by ConspEmoLLM \cite{liu2024conspemollm3}. The figures show that most fake information convey more negative sentiments/emotions and less positive emotions compared to real categories\footnote{Rumours are a complex category, encompassing true rumours, false rumours, and unverified rumours. Due to space constraints, a detailed analysis is not provided here. Nevertheless, different types convey different affective information.}. Figure \ref{fig:3DvisualizationAMT1} and Figure \ref{fig:3DvisualizationPHEME1} present the 3D visualization of affective embeddings on AMTCele and PHEME respectively. Table \ref{tab:statisticsvaluesreg2} shows the statistics values of EIreg and Vreg on PHEME and COCO. 

To explore the relationship between affective classification information and misinformation, we conduct a chi-squared significance test and create two categorical variables. One is the misinformation label (real and fake), and the other variable is affective information. For EIoc, we count the values for 0 (absence) and others (presence) of a certain emotion. For Voc, we count the values of 7 classes. For Ec, we count the number of instances that contain each of the 11 emotions individually. Assuming the null hypothesis that affective signals are independent of text truthfulness, the chi-squared test results in Table \ref{tab:statisticschitest} show p-values close to 0, allowing us to reject the null hypothesis. Overall, affective classification signals are also statistically linked to the veracity of the news.

Table \ref{tab:statisticsvaluesEmbeddings2} shows statistics of different affective embeddings (i.e. last hidden layer of EmoLLaMA-chat-7B). We perform t-tests on the top-$K$ cosine similarity within categories and the cosine similarity between categories. For example, ``fake-legit'' denotes computing the cosine similarity between each data point in the ``fake'' category and each data point in the ``legit'' category. We then selected the top-$K$ similarity values and performed t-test on them. The t-value and p-value of the top-4 similarity values between ``fake-legit'' and ``fake-fake'' are -22.516 and 0, which demonstrates that the top 4 similar data retrieved based on cosine similarity within the ``fake'' category are highly likely to belong to the same ``fake'' category. We can see from the results in Table \ref{tab:statisticsvaluesEmbeddings2} that all affective information leads to the same conclusion in the top-4 scenarios\footnote{It should be noted that in Vreg, as the value of $K$ increases, the second p-value in AMTCele and the first p-value in COCO dataset also gradually increase, which may affect the results. Therefore, we choose $K$ to be 4. The analysis of different values of $K$ can be found in Section \ref{subsec:ablation}.}. We also visualize the data distribution reduced to 3 dimensions using PCA in Figures \ref{fig:3DvisualizationAMT1} and \ref{fig:3DvisualizationPHEME1} in Appendix. It can be observed that different categories are clearly separated in the latent space. All the above demonstrate the close relationship between affective information and misinformation.

\begin{table*}[]
\footnotesize
\resizebox{\textwidth}{!}{
\begin{tabular}{lccccccccc}
\hline
{\color[HTML]{212121} } & \multicolumn{3}{c}{AMT}        & \multicolumn{3}{c}{PHEME}       & \multicolumn{3}{c}{COCO}       \\
                        & EIoc     & Voc      & Ec       & EIoc     & Voc      & Ec        & EIoc     & Voc      & Ec       \\ \hline
chi-squared statistic   & 131.16   & 46.07    & 69.40    & 197.98   & 146.14   & 499.48    & 76.31    & 25.09    & 61.50    \\
p-value                 & 3.60E-25 & 2.86E-08 & 5.78E-11 & 3.08E-39 & 5.07E-29 & 5.69E-101 & 7.76E-14 & 3.28E-04 & 1.88E-09 \\ \hline
\end{tabular}
}
\caption{\label{tab:statisticschitest}
Chi-squared statistics values of EIoc, Voc, Ec on different datasets.}
\end{table*}

\begin{table*}[htb]
\small
\centering
\begin{tabular}{ccccccccc}
\hline
\multirow{2}{*}{Datasets} & \multirow{2}{*}{Affective} & \multirow{2}{*}{sub-emotion} & \multicolumn{2}{c}{non-rumours/related} & \multicolumn{2}{c}{rumours/conspiracy} & \multicolumn{2}{c}{t-test} \\
                          &                            &                              & mean                  & var                   & mean                 & var                  & t            & p           \\ \hline
\multirow{5}{*}{PHEME}    & \multirow{4}{*}{EIreg}     & Anger                        & 0.4547                & 0.0102                & 0.4233               & 0.0075               & 12.7093      & 1.44E-36    \\
                          &                            & Fear                         & 0.5337                & 0.0170                & 0.5632               & 0.0198               & -8.5027      & 2.28E-17    \\
                          &                            & Joy                          & 0.2134                & 0.0121                & 0.1817               & 0.0133               & 11.0177      & 5.58E-28    \\
                          &                            & Sadness                      & 0.5215                & 0.0152                & 0.5177               & 0.0182               & 1.1442       & 0.2526      \\
                          & Vreg                       & \multicolumn{1}{l}{}         & 0.4331                & 0.0143                & 0.3842               & 0.0139               & 15.9786      & 2.18E-56    \\ \hline
\multirow{5}{*}{COCO}     & \multirow{4}{*}{EIreg}     & Anger                        & 0.5475                & 0.0088                & 0.5641               & 0.0068               & -4.5211      & 6.43E-06    \\
                          &                            & Fear                         & 0.5623                & 0.0097                & 0.6034               & 0.0077               & -10.5568     & 1.56E-25    \\
                          &                            & Joy                          & 0.1800                & 0.0111                & 0.1514               & 0.0075               & 7.2230       & 6.66E-13    \\
                          &                            & Sadness                      & 0.4701                & 0.0098                & 0.4773               & 0.0073               & -1.8808      & 0.0601      \\
                          & Vreg                       & \multicolumn{1}{l}{}         & 0.3961                & 0.0095                & 0.3973               & 0.0066               & -0.3325      & 0.7395      \\ \hline
\end{tabular}
\caption{\label{tab:statisticsvaluesreg2}
T-test statistics values of EIreg and Vreg on different datasets. The t-test is conducted between \textit{non-rumours/related} and \textit{rumours/conspiracy}.}
\end{table*}

\begin{table*}[htb]
\resizebox{\textwidth}{!}{
\begin{tabular}{clccccccccccccccc}
\hline
\multicolumn{1}{l}{}         &             & \multicolumn{5}{c}{Vreg}                        & Voc     & Ec      & \multicolumn{4}{c}{EIreg}             & \multicolumn{4}{c}{EIoc}              \\
\multicolumn{1}{l}{Datasets} & Values      & top 4   & top 8   & top 16  & top 32  & top 64  &         &         & anger   & fear    & joy     & sadness & anger   & fear    & joy     & sadness \\ \hline
\multirow{8}{*}{AMTCele}         & fake-legit  & 0.791   & 0.771   & 0.753   & 0.736   & 0.718   & 0.852   & 0.812   & 0.801   & 0.801   & 0.801   & 0.801   & 0.840   & 0.840   & 0.840   & 0.840   \\
                             & fake-fake   & 0.848   & 0.810   & 0.783   & 0.761   & 0.741   & 0.894   & 0.862   & 0.855   & 0.855   & 0.855   & 0.855   & 0.885   & 0.885   & 0.885   & 0.885   \\
                             & t           & -22.516 & -14.875 & -10.951 & -8.976  & -8.037  & -20.550 & -22.617 & -22.434 & -22.433 & -22.462 & -22.461 & -22.260 & -22.246 & -22.267 & -22.244 \\
                             & p           & 0.000   & 0.000   & 0.000   & 0.000   & 0.000   & 0.000   & 0.000   & 0.000   & 0.000   & 0.000   & 0.000   & 0.000   & 0.000   & 0.000   & 0.000   \\
                             & legit-fake  & 0.787   & 0.765   & 0.747   & 0.729   & 0.711   & 0.848   & 0.807   & 0.797   & 0.797   & 0.797   & 0.797   & 0.836   & 0.836   & 0.836   & 0.836   \\
                             & legit-legit & 0.841   & 0.798   & 0.768   & 0.743   & 0.721   & 0.886   & 0.856   & 0.848   & 0.848   & 0.848   & 0.848   & 0.877   & 0.877   & 0.877   & 0.877   \\
                             & t           & -21.568 & -12.845 & -8.052  & -5.263  & -3.452  & -17.138 & -21.024 & -21.399 & -21.387 & -21.407 & -21.396 & -19.364 & -19.328 & -19.335 & -19.315 \\
                             & p           & 0.000   & 0.000   & 0.001   & 0.008   & 0.063   & 0.000   & 0.000   & 0.000   & 0.000   & 0.000   & 0.000   & 0.000   & 0.000   & 0.000   & 0.000   \\ \hline
\multirow{8}{*}{PHEME}       & nonr-rum    & 0.930   & 0.927   & 0.924   & 0.921   & 0.917   & 0.982   & 0.952   & 0.940   & 0.940   & 0.940   & 0.939   & 0.972   & 0.972   & 0.972   & 0.972   \\
                             & nonr-nonr   & 0.957   & 0.946   & 0.938   & 0.932   & 0.927   & 0.989   & 0.971   & 0.963   & 0.963   & 0.963   & 0.963   & 0.983   & 0.983   & 0.983   & 0.983   \\
                             & t           & -75.127 & -49.017 & -35.035 & -27.844 & -24.327 & -69.237 & -78.344 & -77.082 & -77.231 & -76.869 & -78.103 & -71.392 & -71.732 & -71.005 & -72.538 \\
                             & p           & 0.000   & 0.000   & 0.000   & 0.000   & 0.000   & 0.000   & 0.000   & 0.000   & 0.000   & 0.000   & 0.000   & 0.000   & 0.000   & 0.000   & 0.000   \\
                             & rum-nonr    & 0.935   & 0.932   & 0.929   & 0.925   & 0.921   & 0.984   & 0.957   & 0.945   & 0.944   & 0.945   & 0.944   & 0.974   & 0.974   & 0.974   & 0.974   \\
                             & rum-rum     & 0.961   & 0.950   & 0.942   & 0.935   & 0.928   & 0.990   & 0.974   & 0.966   & 0.966   & 0.967   & 0.966   & 0.984   & 0.984   & 0.984   & 0.984   \\
                             & t           & -58.813 & -38.823 & -27.206 & -19.693 & -14.156 & -54.654 & -58.600 & -59.494 & -59.637 & -59.377 & -60.266 & -55.874 & -56.306 & -56.033 & -56.759 \\
                             & p           & 0.000   & 0.000   & 0.000   & 0.000   & 0.000   & 0.000   & 0.000   & 0.000   & 0.000   & 0.000   & 0.000   & 0.000   & 0.000   & 0.000   & 0.000   \\ \hline
\multirow{8}{*}{COCO}        & rela-consp  & 0.873   & 0.870   & 0.866   & 0.861   & 0.856   & 0.955   & 0.905   & 0.885   & 0.885   & 0.886   & 0.885   & 0.936   & 0.936   & 0.937   & 0.936   \\
                             & rela-rela   & 0.907   & 0.887   & 0.875   & 0.865   & 0.857   & 0.967   & 0.931   & 0.916   & 0.916   & 0.916   & 0.916   & 0.953   & 0.953   & 0.954   & 0.954   \\
                             & t           & -44.603 & -23.007 & -11.581 & -5.437  & -2.012  & -37.288 & -43.522 & -44.744 & -44.772 & -44.253 & -44.800 & -38.201 & -38.337 & -37.684 & -38.281 \\
                             & p           & 0.000   & 0.093   & 0.428   & 0.457   & 0.312   & 0.004   & 0.000   & 0.000   & 0.000   & 0.001   & 0.000   & 0.001   & 0.001   & 0.002   & 0.002   \\
                             & consp-rela  & 0.863   & 0.858   & 0.852   & 0.846   & 0.838   & 0.950   & 0.897   & 0.876   & 0.876   & 0.877   & 0.876   & 0.929   & 0.929   & 0.930   & 0.929   \\
                             & consp-consp & 0.911   & 0.891   & 0.878   & 0.868   & 0.859   & 0.968   & 0.933   & 0.919   & 0.919   & 0.920   & 0.920   & 0.954   & 0.954   & 0.955   & 0.954   \\
                             & t           & -74.176 & -47.239 & -33.132 & -25.606 & -21.079 & -54.114 & -69.563 & -73.828 & -73.876 & -73.190 & -73.709 & -60.255 & -60.393 & -59.577 & -60.204 \\
                             & p           & 0.000   & 0.000   & 0.000   & 0.000   & 0.000   & 0.000   & 0.000   & 0.000   & 0.000   & 0.000   & 0.000   & 0.000   & 0.000   & 0.000   & 0.000   \\ \hline
\end{tabular}
}
\caption{\label{tab:statisticsvaluesEmbeddings2}
Statistics values of cosine similarity between embeddings of different affective information on three datasets. Top K denotes retrieval top K examples. In addition to Vreg, the results of other affective information are all based on top 4. ``A-B'' represents the calculation of cosine similarity between each data point in A and each data point in B. Each element (i, j) in the resulting calculation represents the cosine similarity between the i-th vector in the A group embeddings and the j-th vector in the B group embeddings. The top 4 refers to selecting the four highest values from each row. The t-value and p-value represent the t-test results for the ``A-B'' results of the two lines above.}
\end{table*}

\end{document}